\documentclass[10pt,twocolumn,letterpaper]{article}

\usepackage[preprint]{cvpr}      %

\usepackage[dvipsnames]{xcolor}

\definecolor{cvprblue}{rgb}{0.21,0.49,0.74}
\usepackage[pagebackref,breaklinks,colorlinks,citecolor=cvprblue]{hyperref}
\usepackage{booktabs} %
\usepackage{tikz}
\usepackage{pgfplots}
\usetikzlibrary{positioning}
\usetikzlibrary{calc}
\usepackage{graphicx}
\usepackage{subcaption}
\usepackage{etoolbox,lipsum}
\usepackage{xcolor}
\usepackage{afterpage}
\definecolor{lightred}{RGB}{255,200,200}
\definecolor{lightorange}{RGB}{255,215,200}
\definecolor{lightyellow}{RGB}{255,255,200}
\definecolor{skyblue}{rgb}{0.53, 0.81, 0.98}
\usepackage{colortbl}
\usepackage{etoolbox,lipsum}

\def\paperID{82} %
\def\confName{3DV\xspace}
\def\confYear{2025\xspace}

\title{WaterSplatting: Fast Underwater 3D Scene Reconstruction \\Using Gaussian Splatting}

\author{
\textbf{Huapeng Li}$^1$, 
\textbf{Wenxuan Song}$^2$,
\textbf{Tianao Xu}$^2$,
\textbf{Alexandre Elsig}$^2$,
\textbf{Jonas Kulhanek}$^{3,2}$\\
$^1$ University of Zurich;\quad $^2$ ETH Zurich,\quad $^3$ CTU in Prague
}

\newread\imgstream
\immediate\openin\imgstream=imagedata.in
\makeatletter
\def\new@kvginclip#1{}
\def\new@kvgintrim#1{}
\let\old@kvginclip\KV@Gin@clip
\let\old@kvgintrim\KV@Gin@trim
\let\oldincludegraphics\includegraphics
\providecommand{\includegraphics}{}
\renewcommand{\includegraphics}[2][]{%
  \immediate\read\imgstream to \src
  \immediate\read\imgstream to \removecrop
  \ifnum\removecrop=1
      \let\KV@Gin@clip\new@kvginclip
      \let\KV@Gin@trim\new@kvgintrim
  \fi
  \oldincludegraphics[#1]{\src}%
  \let\KV@Gin@clip\old@kvginclip
  \let\KV@Gin@trim\old@kvgintrim}
\makeatother

\begin{document}

\twocolumn[{%
\renewcommand\twocolumn[1][]{#1}%
\maketitle
\vspace{-2.5em}
\begin{center}
\centering
    \begin{tikzpicture}
    \tikzset{%
        img/.style={
            alias=img,
            anchor=south west,
            inner sep=0pt,
            outer sep=0.3mm
        },
        label/.style={
            anchor=north,
            fill=skyblue!50,
            text opacity=1,
            align=center,
            yshift=0.3mm,
            inner sep=0pt,
            outer sep=0pt,
            minimum width=0.163\linewidth,
            minimum height=0.0489\linewidth
        }
    }
        \node[img] at (0,0)
            {\includegraphics[width=.163\linewidth, height=0.10866\linewidth]{figures/front_page/iui3/3dgs.png}};
        \node [label] (lab00) at (img.south) {\fontsize{7}{4}\selectfont 3DGS (451.4 fps)\\\fontsize{7}{4}\selectfont Train: 12min, PSNR: 22.9};
        
        \node[img] at (img.south east)
            {\includegraphics[width=.163\linewidth, height=0.10866\linewidth]{new_figures/front_page/iui3/seathru.png}};
        \node [label] at (img.south) {\fontsize{7}{4}\selectfont SeaThru-NeRF (0.07fps)\\\fontsize{7}{4}\selectfont Train: 10h, PSNR: 25.9};
        
        \node[img] at (img.south east)
            {\includegraphics[width=.163\linewidth, height=0.10866\linewidth]{figures/front_page/iui3/seathru_clear.png}};
        \node [label] at (img.south)  {\fontsize{7}{4}\selectfont SeaThru-NeRF's\\\fontsize{7}{4}\selectfont Restoration};
        
        \node[img] at (img.south east)
            {\includegraphics[width=.163\linewidth, height=0.10866\linewidth]{new_figures/front_page/iui3/ours.png}};
        \node [label] at (img.south)  {\fontsize{7}{4}\selectfont Ours (56.7 fps)\\\fontsize{7}{4}\selectfont Train: 7.2min, PSNR: 30.4};
        
        \node[img] at (img.south east)
            {\includegraphics[width=.163\linewidth, height=0.10866\linewidth]{new_figures/front_page/iui3/ours_clear.png}};
        \node [label] at (img.south)  {\fontsize{7}{4}\selectfont Our Restoration};
        
        \node[img] at (img.south east)
            {\includegraphics[width=.163\linewidth, height=0.10866\linewidth]{new_figures/front_page/iui3/gt.png}};
        \node [label] at (img.south) {\fontsize{7}{4}\selectfont Ground Truth};
        
        \node[img,anchor=north,yshift=-0.3mm] at (lab00.south)
            {\includegraphics[width=.163\linewidth, height=0.10866\linewidth]{figures/front_page/japan/3dgs.png}};
        \node [label] at (img.south) {\fontsize{7}{4}\selectfont 3DGS (424.0 fps)\\\fontsize{7}{4}\selectfont Train: 13min, PSNR: 21.5};
        
        \node[img] at (img.south east)
            {\includegraphics[width=.163\linewidth, height=0.10866\linewidth]{new_figures/front_page/japan/seathru.png}};
        \node [label] at (img.south)  {\fontsize{7}{4}\selectfont SeaThru-NeRF (0.07fps)\\\fontsize{7}{4}\selectfont Train: 10h, PSNR: 21.8};
    
        \node[img] at (img.south east)
            {\includegraphics[width=.163\linewidth, height=0.10866\linewidth]{figures/front_page/japan/seathru_clear.png}};
        \node [label] at (img.south) {\fontsize{7}{4}\selectfont SeaThru-NeRF's\\\fontsize{7}{4}\selectfont Restoration};
        
        \node[img] at (img.south east)
            {\includegraphics[width=.163\linewidth, height=0.10866\linewidth]{new_figures/front_page/japan/ours.png}};
        \node [label] at (img.south) {\fontsize{7}{4}\selectfont  Ours (56.9 fps)\\\fontsize{7}{4}\selectfont Train: 7.7min, PSNR: 24.9};
        
        \node[img] at (img.south east)
            {\includegraphics[width=.163\linewidth, height=0.10866\linewidth]{new_figures/front_page/japan/ours_clear.png}};
        \node [label] at (img.south)  {\fontsize{7}{4}\selectfont Our Restoration};
        
        \node[img] at (img.south east)
            {\includegraphics[width=.163\linewidth, height=0.10866\linewidth]{new_figures/front_page/japan/gt.png}};
        \node [label] at (img.south) {\fontsize{7}{4}\selectfont Ground Truth};
    \end{tikzpicture}
    \vspace{-0.5em}
    
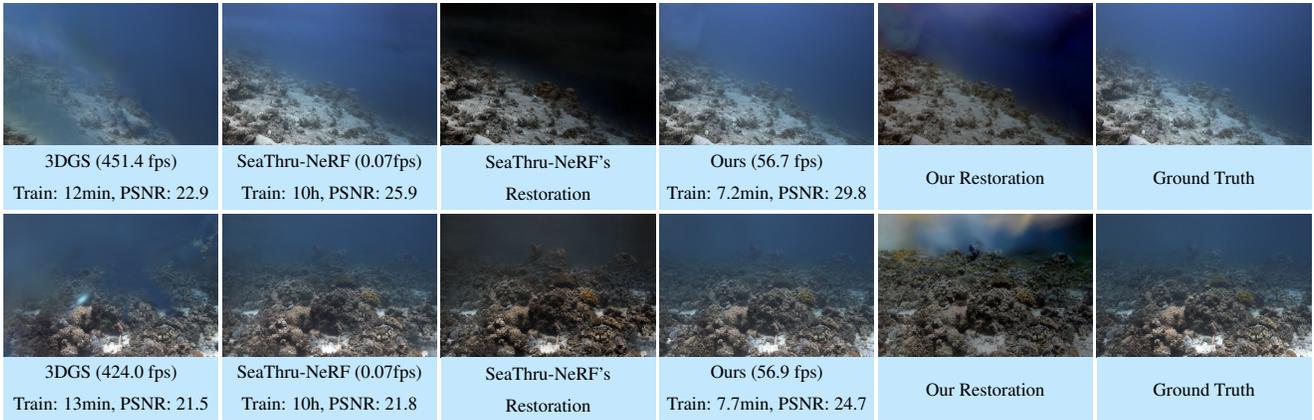
\captionof{figure}{Our approach surpasses the performance of state-of-the-art NeRF-based underwater reconstruction methods \cite{levy2023seathrunerf} while offering real-time rendering speed \cite{kerbl3Dgaussians}.}
    \label{fig:frontimg}
\end{center}
}]

\begin{abstract}
The underwater 3D scene reconstruction is a challenging, yet interesting problem with applications ranging from naval robots to VR experiences.
The problem was successfully tackled by fully volumetric NeRF-based methods which can model both the geometry and the medium (water). 
Unfortunately, these methods are slow to train and do not offer real-time rendering. More recently, 3D Gaussian Splatting (3DGS) method offered a fast alternative to NeRFs. However, because it is an explicit method that renders only the geometry, it cannot render the medium and is therefore unsuited for underwater reconstruction. Therefore, we propose a novel approach that fuses volumetric rendering with 3DGS to handle underwater data effectively. Our method employs 3DGS for explicit geometry representation and a separate volumetric field (queried once per pixel) for capturing the scattering medium. This dual representation further allows the restoration of the scenes by removing the scattering medium. Our method outperforms state-of-the-art NeRF-based methods in rendering quality on the underwater SeaThru-NeRF dataset. Furthermore, it does so while offering real-time rendering performance, addressing the efficiency limitations of existing methods.
\\\textbf{Web:} \texttt{\url{https://water-splatting.github.io}}
\end{abstract} 
\section{Introduction}
Neural Radiance Fields (NeRFs) \cite{mildenhall2020nerf} have recently gained significant popularity due to their ability to offer photorealistic 3D scene reconstruction quality. This has opened up new avenues in the field of 3D rendering and reconstruction. However, the landscape of 3D rendering techniques is rapidly evolving. More recently, point splatting methods have experienced a resurgence in the form of 3D Gaussian Splatting (3DGS) \cite{kerbl3Dgaussians}, which matches NeRFs in terms of rendering quality and offers real-time rendering speed, better editability, and control.

The reconstruction of scattering scenes, such as foggy and underwater environments, is an interesting research area with applications ranging from naval robots to VR experiences. Reconstructing geometry inside a water volume is challenging due to the presence of the scattering medium with properties different from air. In a typical scene, the primary requirement is to represent the surface. Both NeRFs and Gaussian splatting methods are optimized to focus on representing the surfaces only, thereby gaining better efficiency. In the case of NeRFs, since they are fully volumetric, they should theoretically be able to represent the medium fully volumetrically. However, this is no longer the case as the proposal sampler used to speed up NeRFs prevents them from learning volumes well.

To address this issue, a NeRF approach, SeaThru-NeRF \cite{levy2023seathrunerf}, was proposed, which uses two fields: one for the geometry and one for the volume in between. However, it is slow in both rendering and training. Therefore, we propose a novel approach to represent the geometry explicitly using 3DGS but to represent the volume in between using a volumetric representation. The renderer we propose not only surpasses the rendering quality of fully volumetric representations, as demonstrated by \cite{levy2023seathrunerf} but also achieves rendering and training speeds comparable to 3DGS.

To validate our method, we evaluate it on the established benchmark underwater dataset - SeaThru-NeRF \cite{levy2023seathrunerf}. The results of our evaluation demonstrate the effectiveness of our proposed method in achieving high-quality, efficient underwater reconstruction.
In summary, we make the following contributions:

1. \textbf{Splatting with Medium:} We introduce a novel approach that combines the strengths of Gaussian Splatting (GS) and volume rendering. Our method employs GS for explicit geometry representation and a separate volumetric field for capturing the scattering medium. This dual representation allows for the synthesis of novel views in scattering media and the restoration of clear scenes without a medium. 

2. \textbf{Loss Function Alignment:} We propose a novel loss function designed to align 3DGS with human perception of High Dynamic Range (HDR) and low-light scenes. 

3. \textbf{Efficient Synthesis and Restoration:} We demonstrate that our method outperforms other models on synthesizing novel view on real-world underwater data and restoring clean scenes on synthesized back-scattering scenes with much shorter training and rendering time.

\section{Related Work}
\subsection{NeRF}
The field of 3D scene reconstruction has gained significant attention with the advent of NeRF methods \cite{mildenhall2020nerf,Lombardi:2019,SunSC22}. NeRFs represent the 3D scene as a radiance field—comprising differential volume density and view-dependent color—rendered using a volume rendering integral from a list of samples sampled along the ray \cite{queianchen_nerf}. Originally, NeRFs utilized Multilayer Perceptrons (MLPs) for representing the radiance field \cite{mildenhall2020nerf,barron2021mipnerf,barron2022mipnerf360}, but they were slow to train and render. To accelerate the training and rendering processes, alternative methods have been proposed using discrete grids \cite{yu2021plenoctrees,yu2022plenoxels}, hash grids \cite{mueller2022instant,nerfstudio,barron2023zipnerf}, tensorial decomposition \cite{Chen2022ECCV,kplanes_2023}, point clouds \cite{xu2022point}, or tetrahedral mesh \cite{kulhanek2023tetranerf}. NeRFs have been enhanced in various ways, including improved anti-aliasing \cite{barron2021mipnerf,barron2023zipnerf}, handling of large 3D scenes \cite{Tancik_2022_CVPR}, and complex camera trajectories \cite{barron2022mipnerf360,wang2023f2nerf}. Moreover, NeRFs have been extended to a wide range of applications such as semantic segmentation \cite{cen2023segment,garfield2024}, few-view novel view synthesis \cite{chan2023genvs,chen2021mvsnerf,yu2021pixelnerf,wu2023reconfusion,lin2020sdfsrn}, and generative 3D modeling \cite{poole2022dreamfusion,liu2023zero1to3}. Despite these advancements, the slow rendering speed of NeRFs remains a critical limitation, hindering their widespread adoption on end-user devices.

\subsection{3D Gaussian Splatting}

Recently, Gaussian Splatting (3DGS) \cite{kerbl3Dgaussians} has seen a resurgence as a powerful method for real-time 3D rendering, matching the quality of Neural Radiance Fields (NeRFs) \cite{mildenhall2020nerf} but with significantly faster speeds even suitable for end-user devices \cite{kerbl3Dgaussians}. This technique enhances control and editability since scenes are stored as editable sets of Gaussians, allowing for modifications, merging, and other manipulations. Additionally, the original 3DGS method has been refined to improve anti-aliasing \cite{Yu2023MipSplatting} and adapt density control more effectively \cite{ye2024absgs}. Owing to these advancements, 3DGS has been widely adopted in various applications such as large-scale reconstructions \cite{lin2024vastgaussian}, 3D generation \cite{chung2023luciddreamer}, simultaneous localization and mapping (SLAM) \cite{li2024sgsslam, keetha2023splatam}, and open-set segmentation \cite{qin2023langsplat}. Despite its state-of-the-art rendering quality and impressive handling of complex scenes, 3DGS's explicit representation nature limits its use in scenarios requiring the depiction of semi-transparent volumes, such as underwater reconstructions where light scattering and absorption are significant challenges \cite{kerbl3Dgaussians}.

\subsection{Computer Vision in Scattering Media}
There are many challanges in underwater computer vision.
The complex lighting conditions including scattering and attenuation of light leading to distorted images and the failure of traditional algorithms trained on clear-air scenes \cite{10.1145/3578516}.
SeaThru \cite{akkaynak2019removing}
introduces a method for removing water from underwater images. This method addresses color distortion in underwater images by revising the image formation model from \cite{akkaynak2018revised}, accurately estimating backscatter, and correcting colors along the depth axis. 
WaterNeRF\cite{sethuraman2023waternerf} estimates medium parameters separately from rendering, while ScatterNeRF\cite{ramazzina2023scatternerf} extends NeRF's volumetric rendering to model scattering in adverse weather.
\cite{zhang2023beyond} introduces a neural reflectance field to jointly learn scene albedo, normals, and medium effects, achieving color consistency by modeling light attenuation and backscatter through logistic regression and differentiable volume rendering.
\cite{tang2024uwnerf} explores hybrid neural-explicit approaches for reconstructing dynamic underwater environments, though its reliance on volumetric rendering limits real-time applicability.
SeaThru-NeRF \cite{levy2023seathrunerf} implements the image formation model \cite{akkaynak2018revised} that separates direct and backscatter components, into the NeRF rendering equations, which is highly specialized for underwater scenes. We implemented a similar model but on Gaussian Splatting, which yields higher performance and enables real-time rendering.
\section{Method}
\begin{figure*}[ht!]
    \centering
    \begin{subfigure}[b]{1\linewidth}
        \includegraphics[width=\linewidth]{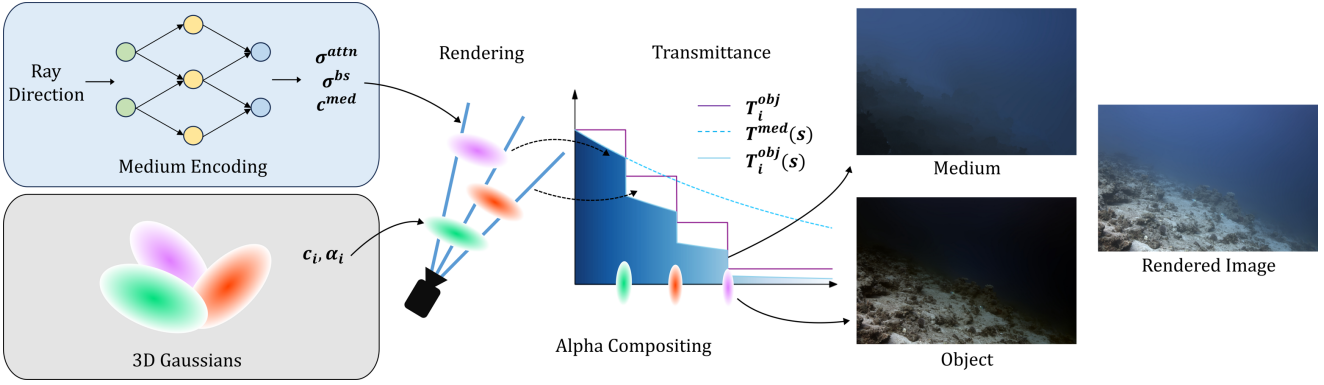}
        
    \end{subfigure}
\caption{\textbf{Splatting with Medium}: We start rendering by casting a ray per pixel and collect the patch-intersected Gaussians along the ray and their color given ray direction. Then, we walk through the list of sorted Gaussians per pixel, query their opacity and depth, based on which we acquire the transmittance of both Gaussians and medium, rendering the Gaussians and the segments between adjacent two to obtain the \textbf{Medium} component and the \textbf{Object} component.}
\label{fig:pipeline}
\end{figure*}
We start by briefly reviewing 3DGS and the rendering model in scattering media in Sec. \ref{subsec:Preliminaries}. Then, we illustrate our proposed rendering model combining 3DGS with medium encoding in Sec. \ref{subsec:Splatting with Medium}. At last, we explain our proposed loss function to align 3DGS with human perception of HDR scenes in Sec. \ref{subsec:Loss Function Alignment}.

\subsection{Preliminaries}
\label{subsec:Preliminaries}
\textbf{3D Gaussian Splatting} models a scene with explicit learnable primitives {$\mathcal{G}_0$,  $\mathcal{G}_1$, ..., $\mathcal{G}_N$}. Each Gaussian $\mathcal{G}_i$ is defined by a central position $\mu_i$ and covariance matrix $\Sigma_i$ \cite{kerbl3Dgaussians}:
\begin{equation}
    G_i(p)=e^{-\frac{1}{2}(p-\mu_i)^T(\Sigma_i)^{-1}(p-\mu_i)}.
    \label{eq:3d-density}
\end{equation}
3DGS primitive also has two additional parameterized properties: opacity $o_i$ and spherical harmonics coefficients $SH_i$ to represent directional appearance component (anisotropic color). In order to render pixel-wise color, primitives are transformed into camera space via a viewing transformation $W$, the Jacobian of the affine approximation of the projective transformation $J$ on $\Sigma_i$ and a projection matrix $P$. Then we get the projected 2D means $\hat\mu_i$ and 2D covariance matrices $\hat\Sigma_i$:
\begin{equation}
    \hat{\Sigma}_i= (J W \Sigma_i W^T J^T)_{1:2,1:2}\,,\quad
    \hat{\mu}_i = (P\mu_i)_{1:2}\,,
    \label{eq:2d-cov}
\end{equation}
and the depth of $\mathcal{G}_i$ on z-coordinate:
\begin{equation}
    s_i = (P\mu_i)_3.
    \label{eq:depth}
\end{equation}
The Gaussian kernel $\hat{G}_i$ of 2D Gaussian is represented as:
\begin{equation}
    \hat{G}_i(p)=e^{-\frac{1}{2}(p-\hat{\mu}_i)^T(\hat{\Sigma}_i)^{-1}(p-\hat{\mu}_i)},
    \label{eq:2d-density}
\end{equation}
where $p$ is the coordinate of the pixel. For rasterization, each Gaussian is truncated at 3 sigma, considering only those intersecting with the patch comprising $16 \times 16$ pixels within this range, based on the property that about $99.7\%$ of the probability lies within 3 sigma of the mean. Pixel colors are computed by alpha blending of the sorted intersected Gaussians $\mathcal{G}_i$ whose $\alpha_i$ are higher than a threshold:
\begin{equation}
    C = \sum_{i=1}^N c_i \alpha_i\prod_{j=1}^{i-1}(1-\alpha_j)\,,\quad
    \alpha_i=\sigma(o_i)\cdot \hat{G_i}(p)\,,
    \label{eq:gs-alpha}
\end{equation}
where $c_i$ is the color given the view direction, $\sigma(\cdot)$ is the Sigmoid function and N is the number of Gaussians involved in alpha blending.
During optimization, 3DGS periodically densify Gaussians with high average gradient on 2D coordinates $\hat{\mu}_i$ across frames via splitting large ones and duplicating small ones. In the meantime, 3DGS prunes primitives with low opacity for acceleration and periodically set $\alpha_i$ close to zero for all Gaussians to moderate the increase of floaters close to the input cameras.

For \textbf{scene rendering in scattering media} we use the revised underwater image formation model from \cite{akkaynak2018revised} where the final image I is separated into a direct and backscatter component
\begin{equation}
    I = \underbrace{ O \cdot e^{-\beta^D(\mathbf{v}_D)\cdot z}}_{\text{Direct Image component}} + \underbrace{B^\infty \cdot (1-e^{-\beta^B(\mathbf{v}_B)\cdot z})}_{\text{Backscatter Image component}},
    \label{eq:revised}
\end{equation}
where $O$ is the clear scene captured at depth $z$ in no medium, $B^\infty$ is the backscatter color of the water at infinite distance. The colors get multiplied with attenuations, where the $\beta^D$ and $\beta^B$ are attenuation coefficients for the direct and backscatter components of the image which represent the effect the medium has on the color. The vector $\mathbf{v}^D$ represents the dependencies for the direct component, which includes factors such as the depth $z$, reflectance, ambient light, water scattering properties, and the attenuation coefficient of the water. The vector $\mathbf{v}^B$ represents the dependencies for the backscatter component, which includes ambient light, water scattering properties, the backscatter coefficient, and the attenuation coefficient of the water.

\subsection{Splatting with Medium}
\label{subsec:Splatting with Medium}
We illustrate the pipeline of our method in Fig. \ref{fig:pipeline}. The input to our model is a set of images with scattering medium and corresponding camera poses. We initialize a set of 3D Gaussians via SfM \cite{kerbl3Dgaussians} and optimize them with medium properties encoded by a neural network. Under the occlusion of both primitives and medium, our model acquires the transmittance along the ray and is capable of synthesizing medium component and object component in the novel view. Below we derive the whole model in detail.

Considering the expected color of a pixel integrated along the camera ray $r(s) = o + d(s)$ from the camera to infinitely far $C(r)=\int_{0}^{\infty} T(s)\sigma(s)c(s)ds$ \cite{kajiya1984ray} because of unbounded rendering of 3DGS \cite{kerbl3Dgaussians}, we release the constraints on light traveling in clear air to through a scattering medium \cite{levy2023seathrunerf} by adding the medium term:
\begin{equation}
    C(r) = \int_{0}^{\infty} T(s) (\sigma^{\text{obj}}(s) c^{\text{obj}}(s) + \sigma^{\text{med}}(s) c^{\text{med}}(s)) ds
    \label{eq:begin-color}
\end{equation}
\begin{equation}
    T(s) = exp(-\int_{0}^{\infty} (\sigma^{\text{obj}}(s) + \sigma^{\text{med}}(s)) ds),
    \label{eq:begin-trans}
\end{equation}
where $\sigma^{\text{obj}}$/$\sigma^{\text{med}}$ and $c^{\text{obj}}$/$c^{\text{med}}$ are density and color of objects and medium respectively.

Following \cite{levy2023seathrunerf}, we take $\sigma^{\text{med}}$ and $c^{\text{med}}$ to be constant per ray and separate per color channel. In order to apply discretized representation in 3DGS, the transmittance $T_i(s)$ in front of the i-th Gaussian $\mathcal{G}_i$ (and behind (i-1)-th Gaussian $\mathcal{G}_{i-1}$) with depth $s\in[s_{i-1}, s_i]$ can be decomposed as
\begin{equation}
    T_i(s) = T_i^{\text{obj}} T^{\text{med}}(s),\quad
    T_i^{\text{obj}} = \prod_{j=1}^{i-1} (1 - \alpha_j)
    \label{eq:trans-decompose}
\end{equation}
where $T_i^{\text{obj}}$
is the accumulated transmittance contributed by previous primitives' occlusion \cite{kerbl3Dgaussians} and
\begin{equation}
        T^{\text{med}}(s) = exp(-\int_{0}^{s} \sigma^{\text{med}}(s) ds)
        =exp(-\sigma^{\text{med}} s)
    \label{eq:trans-med}
\end{equation}
is the accumulated transmittance under the effect of medium from the camera to depth $s$.
Then, the color is composed with discretized Gaussians and integrable medium
\begin{equation}
    C(r) = \sum_{i=1}^{N} C_i^{\text{obj}}(r) + \sum_{i=1}^{N} C_i^{\text{med}}(r).
    \label{eq:inter-color}
\end{equation}
The contributed color of the $\mathcal{G}_i$ to final output is
\begin{equation}
    \begin{split}
        C^{\text{obj}}_i(r)&=T_i^{\text{obj}} T^{\text{med}}(s_i) \alpha_i c_i\\
        &=T_i^{\text{obj}} \alpha_i c_i exp(-\sigma^{\text{med}} s_i),
    \end{split}
    \label{eq:inter-obj}
\end{equation}
where $\alpha_i$ is the opacity given the relative position between the pixel $p$ and $\mu_i$ in Eq. (\ref{eq:gs-alpha}) and $c_i=c_i^{\text{obj}}$ is the color given the ray direction. The contributed color of the medium between the (i-1)-th and $\mathcal{G}_i$ is
\begin{equation}
    \begin{split}
        C_i^{\text{med}}(r)&=\int_{s_{i-1}}^{s_i} T_i^{\text{obj}} T^{\text{med}}(s) \sigma^{\text{med}} c^{\text{med}} ds\\
        &=T_i^{\text{obj}} c^{\text{med}} [exp(-\sigma^{\text{med}} s_{i-1}) - exp(-\sigma^{\text{med}} s_i)].
    \end{split}
    \label{eq:inter-med}
\end{equation}

To precisely estimate the properties of medium, we also include the background medium term from the last Gaussian $\mathcal{G}_N$ to infinitely far
\begin{equation}
    \begin{split}
        C_\infty^{\text{med}}(r)&=\int_{s_{N}}^{\infty} T_i^{\text{obj}} T^{\text{med}}(s) \sigma^{\text{med}} c^{\text{med}} ds\\
        &=T_i^{\text{obj}} c^{\text{med}} exp(-\sigma^{\text{med}} s_{N})
    \end{split}
    \label{eq:inter-inf}
\end{equation}
into the accumulated color. 

As discussed in \cite{akkaynak2018revised}, the effective $\sigma^{\text{med}}$ experienced a camera with wide-band color channels by differs in $C_{\cdot}^{\text{obj}}(r)$ and $C_{\cdot}^{\text{med}}(r)$, following \cite{levy2023seathrunerf}, we use two sets of parameters, object attenuation $\sigma^{attn}$ and medium back-scatter $\sigma^{\text{bs}}$ for $C_i^{\text{obj}}(r)$ and $C_i^{\text{med}}(r)$ respectively.
By setting $s_0=0$, our final equations of rendered color are:
\begin{equation}
    \begin{split}
        C(r) = \sum_{i=1}^{N} C_i^{\text{obj}}(r) + \sum_{i=1}^{N} C_i^{\text{med}}(r) + C_\infty^{\text{med}}(r),
    \end{split}
    \label{eq:final-color}
\end{equation}
\begin{equation}
    \begin{split}
        C_i^{\text{obj}}(r)=T_i^{\text{obj}} \alpha_i c_i exp(-\sigma^{attn} s_i),
    \end{split}
    \label{eq:final-obj}
\end{equation}
\begin{equation}
    \begin{split}
        C_i^{\text{med}}(r)=T_i^{\text{obj}} c^{\text{med}} [exp(-\sigma^{\text{bs}} s_{i-1})) - exp(-\sigma^{\text{bs}}s_i)],
    \end{split}
    \label{eq:final-med}
\end{equation}
\begin{equation}
    \begin{split}
        C_\infty^{\text{med}}(r)=T_N^{\text{obj}} c^{\text{med}} exp(-\sigma^{\text{bs}} s_{N}).
    \end{split}
    \label{eq:final-inf}
\end{equation}

\subsection{Loss Function Alignment}
\label{subsec:Loss Function Alignment}
In vanilla 3DGS, the loss function is a combination of $\mathcal{L}_1$ loss and D-SSIM loss.  %
In low-light situations, \cite{mildenhall2021nerf} proposed a regularized $\mathcal{L}_2$ loss
\begin{equation}
    \mathcal{L}_{\text{Reg-}\mathcal{L}_{2}} = ((sg(\hat{y})+\epsilon)^{-1}\odot(\hat{y}-y))^2,
    \label{eq:l2}
\end{equation}
to boost the weight of the dark regions in optimization to align with how humans perceive dynamic range, which is applied to underwater scene recontruction by \cite{levy2023seathrunerf}. For the case of our 3DGS-based model, we propose a regularized loss function $\mathcal{L_{\text{Reg}}}$: we apply pixel-wise weight $W=\{w_{i,j}\}$ on both rendered estimate $\hat{y}$ and target image $y$, where  $w_{i,j}=(sg(\hat{y}_{i,j})+\epsilon)^{-1}$ with pixel coordinate $(i, j)$ and $sg(\cdot)$ denotes stopping gradient of its argument, which backpropagates zero derivative.

By introducing the weight $W$ to the loss function of 3DGS composing of $\mathcal{L}_1$ and D-SSIM loss,
we have the regularized $\mathcal{L}_1$ loss
\begin{equation}
    \mathcal{L}_{\text{Reg-}\mathcal{L}_{1}} = |W\odot(\hat{y}-y)|,
    \label{eq:l1}
\end{equation}
 and the regularized D-SSIM loss
\begin{equation}
    \mathcal{L}_{\text{Reg-DSSIM}} = \mathcal{L}_{\text{DSSIM}}(W\odot y, W\odot\hat{y}).
    \label{eq:ssim}
\end{equation}

$\mathcal{L}_{\text{Reg-DSSIM}}$ prioritizes structural similarity and perceptual fidelity, particularly in dark regions where human perception is more sensitive. Integrating regularization into the $\mathcal{L}_{\text{Reg-DSSIM}}$ formulation becomes particularly critical for 3DGS optimization due to the discrete nature of its primitives, necessitating structural regularization to maintain perceptual consistency across independently optimized elements - a distinct requirement from NeRF-based approaches where parameter-shared volumetric representations can achieve adequate low-light adaptation through pixel-level regularized losses alone. To model the smoothness of volumetric medium, we employ $\mathcal{L}_{\text{Reg-}\mathcal{L}_{2}}$ as our pixel-level loss.
Our final proposed loss function is
\begin{equation}
    \mathcal{L}_{\text{Reg}}=(1-\lambda)\mathcal{L}_{\text{Reg-}\mathcal{L}_{2}}+\lambda\mathcal{L}_{\text{Reg-DSSIM}} \,.
\end{equation}

\section{Experiments}
\begin{figure*}[ht!]
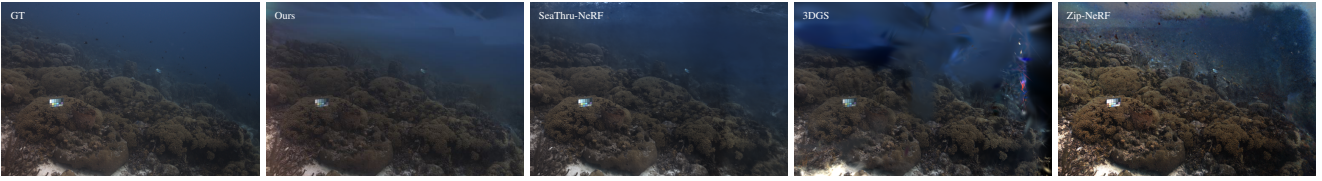

    \centering
    \begin{subfigure}[b]{.196\linewidth}
        \centering
        \begin{tikzpicture}
            \node[anchor=south west, inner sep=0pt] (img) at (0,0)
                {\includegraphics[width=\linewidth, height=0.67\linewidth]{new_figures/curasao_baselines/gt.png}};
            \begin{scope}[x={(img.south east)},y={(img.north west)}]         
            \node [anchor=north west, white, align=left] at (0,1) {\fontsize{4}{4}\selectfont GT};
            \end{scope}
            
        \end{tikzpicture}
    \end{subfigure}
    \begin{subfigure}[b]{.196\linewidth}
        \centering
        \begin{tikzpicture}
            \node[anchor=south west, inner sep=0pt] (img) at (0,0)
                {\includegraphics[width=\linewidth, height=0.67\linewidth]{new_figures/curasao_baselines/ours.png}};
           \begin{scope}[x={(img.south east)},y={(img.north west)}]
        
        \node [anchor=north west, white, align=left] at (0,1) {\fontsize{4}{4}\selectfont Ours};
    \end{scope}
        \end{tikzpicture}
    \end{subfigure}
    \begin{subfigure}[b]{.196\linewidth}
        \centering
        \begin{tikzpicture}
            \node[anchor=south west, inner sep=0pt] (img) at (0,0)
                {\includegraphics[width=\linewidth, height=0.67\linewidth]{new_figures/curasao_baselines/seathru.png}};
            \begin{scope}[x={(img.south east)},y={(img.north west)}]            
            \node [anchor=north west, white, align=left] at (0,1) {\fontsize{4}{4}\selectfont SeaThru-NeRF};
            \end{scope}
        \end{tikzpicture}
    \end{subfigure}
    \begin{subfigure}[b]{.196\linewidth}
        \centering
        \begin{tikzpicture}
            \node[anchor=south west, inner sep=0pt] (img) at (0,0)
                {\includegraphics[width=\linewidth, height=0.67\linewidth]{figures/curasao_baselines/3dgs.png}};
            \begin{scope}[x={(img.south east)},y={(img.north west)}]         
            \node [anchor=north west, white, align=left] at (0,1) {\fontsize{4}{4}\selectfont 3DGS};
            \end{scope}
        \end{tikzpicture}
    \end{subfigure}
    \begin{subfigure}[b]{.196\linewidth}
        \centering
        \begin{tikzpicture}
            \node[anchor=south west, inner sep=0pt] (img) at (0,0)
                {\includegraphics[width=\linewidth, height=0.67\linewidth]{new_figures/curasao_baselines/zipnerf.png}};
            \begin{scope}[x={(img.south east)},y={(img.north west)}]         
            \node [anchor=north west, white, align=left] at (0,1) {\fontsize{4}{4}\selectfont Zip-NeRF};
            \end{scope}
        \end{tikzpicture}
    \end{subfigure}
    
    \caption{Underwater scene rendering in the 'Curasao' scene. From left to right: white-balanced ground-truth image, our result, SeaThru-NeRF's result, 3DGS' result, and Zip-NeRF's result. Both traditional 3DGS and NeRF with a proposal sampler cannot handle semi-transparent medium well.
    }
    \label{fig:curasao_baselines}
\end{figure*}
\begin{figure*}[ht!]
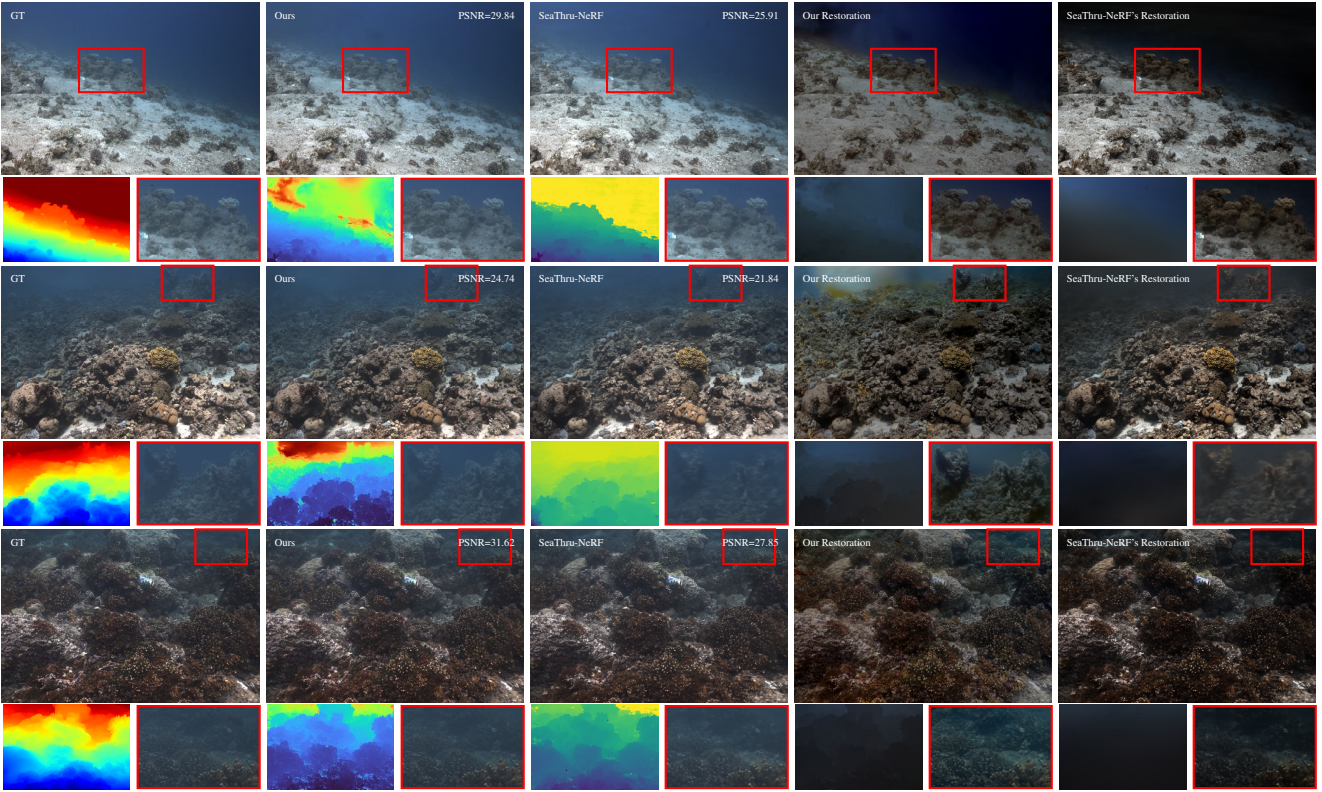

    \centering
    \begin{subfigure}[b]{.196\linewidth}
        \centering
        \begin{tikzpicture}
            \node[anchor=south west, inner sep=0pt] (img) at (0,0)
                {\includegraphics[width=\linewidth, height=0.67\linewidth]{new_figures/red_square/iui/gt.png}};
            \begin{scope}[x={(img.south east)},y={(img.north west)}]         \draw[red,thick] (0.3,0.48) rectangle (0.55,0.73);
            \node [anchor=north west, white, align=left] at (0,1) {\fontsize{4}{4}\selectfont GT};
            \end{scope}
        \end{tikzpicture}
        \begin{minipage}{.49\linewidth}
            \includegraphics[width=\linewidth, height=0.67\linewidth]{new_figures/red_square/iui/depth_iui3_gt.png}
        \end{minipage}
        \begin{minipage}{.47\linewidth}
            \begin{tikzpicture}
                \node[inner sep=0pt] (img) at (0,0)
                    {\includegraphics[trim=415 439.4 612.5 245.6, clip, width=\linewidth, height=0.67\linewidth]{new_figures/red_square/iui/gt.png}};
                \draw[red,thick] (img.south west) rectangle (img.north east); 
            \end{tikzpicture}
        \end{minipage}
    \end{subfigure}
    \begin{subfigure}[b]{.196\linewidth}
        \centering
        \begin{tikzpicture}
            \node[anchor=south west, inner sep=0pt] (img) at (0,0)
                {\includegraphics[width=\linewidth, height=0.67\linewidth]{new_figures/red_square/iui/ours.png}};
           \begin{scope}[x={(img.south east)},y={(img.north west)}]
        \draw[red,thick] (0.3,0.48) rectangle (0.55,0.73);
        \node [anchor=north west, white, align=left] at (0,1) {\fontsize{4}{4}\selectfont Ours};
        \node [anchor=north east, white, align=right] at (1,1) {\fontsize{4}{4}\selectfont PSNR=29.39};
    \end{scope}
        \end{tikzpicture}
        \begin{minipage}{.49\linewidth}
            \includegraphics[width=\linewidth, height=0.67\linewidth]{new_figures/red_square/iui/iui3_depth.png}
        \end{minipage}
        \begin{minipage}{.47\linewidth}
            \begin{tikzpicture}
                \node[inner sep=0pt] (img) at (0,0)
                    {\includegraphics[trim=415 439.4 612.5 245.6, clip, width=\linewidth, height=0.67\linewidth]{new_figures/red_square/iui/ours.png}};
                \draw[red,thick] (img.south west) rectangle (img.north east); 
            \end{tikzpicture}
        \end{minipage}
    \end{subfigure}
    \begin{subfigure}[b]{.196\linewidth}
        \centering
        \begin{tikzpicture}
            \node[anchor=south west, inner sep=0pt] (img) at (0,0)
                {\includegraphics[width=\linewidth, height=0.67\linewidth]{new_figures/red_square/iui/seathru.png}};
            \begin{scope}[x={(img.south east)},y={(img.north west)}]         \draw[red,thick] (0.3,0.48) rectangle (0.55,0.73);     
            \node [anchor=north west, white, align=left] at (0,1) {\fontsize{4}{4}\selectfont SeaThru-NeRF};
            \node [anchor=north east, white, align=right] at (1,1) {\fontsize{4}{4}\selectfont PSNR=27.31};
            \end{scope}
        \end{tikzpicture}
        \begin{minipage}{.49\linewidth}
            \includegraphics[width=\linewidth, height=0.67\linewidth]{figures/red_square/iui3/seathru_depth.png}
        \end{minipage}
        \begin{minipage}{.47\linewidth}
            \begin{tikzpicture}
                \node[inner sep=0pt] (img) at (0,0)
                    {\includegraphics[trim=415 439.4 612.5 245.6, clip, width=\linewidth, height=0.67\linewidth]{new_figures/red_square/iui/seathru.png}};
                \draw[red,thick] (img.south west) rectangle (img.north east); 
            \end{tikzpicture}
        \end{minipage}
    \end{subfigure}
    \begin{subfigure}[b]{.196\linewidth}
        \centering
        \begin{tikzpicture}
            \node[anchor=south west, inner sep=0pt] (img) at (0,0)
                {\includegraphics[width=\linewidth, height=0.67\linewidth]{new_figures/red_square/iui/ours_clear.png}};
            \begin{scope}[x={(img.south east)},y={(img.north west)}]         \draw[red,thick] (0.3,0.48) rectangle (0.55,0.73);     
            \node [anchor=north west, white, align=left] at (0,1) {\fontsize{4}{4}\selectfont Our Restoration};
            \end{scope}
        \end{tikzpicture}
        \begin{minipage}{.49\linewidth}
            \includegraphics[width=\linewidth, height=0.67\linewidth]{new_figures/red_square/iui/ours_medium.png}
        \end{minipage}
        \begin{minipage}{.47\linewidth}
            \begin{tikzpicture}
                \node[inner sep=0pt] (img) at (0,0)
                    {\includegraphics[trim=415 439.4 612.5 245.6, clip, width=\linewidth, height=0.67\linewidth]{new_figures/red_square/iui/ours_clear.png}};
                \draw[red,thick] (img.south west) rectangle (img.north east); 
            \end{tikzpicture}
        \end{minipage}
    \end{subfigure}
    \begin{subfigure}[b]{.196\linewidth}
        \centering
        \begin{tikzpicture}
            \node[anchor=south west, inner sep=0pt] (img) at (0,0)
                {\includegraphics[width=\linewidth, height=0.67\linewidth]{figures/red_square/iui3/seathru_clear.png}};
            \begin{scope}[x={(img.south east)},y={(img.north west)}]         \draw[red,thick] (0.3,0.48) rectangle (0.55,0.73);     
            \node [anchor=north west, white, align=left] at (0,1) {\fontsize{4}{4}\selectfont SeaThru-NeRF's Restoration};
            \end{scope}
        \end{tikzpicture}
        \begin{minipage}{.49\linewidth}
            \includegraphics[width=\linewidth, height=0.67\linewidth]{figures/red_square/iui3/seathru_medium.png}
        \end{minipage}
        \begin{minipage}{.47\linewidth}
            \begin{tikzpicture}
                \node[inner sep=0pt] (img) at (0,0)
                    {\includegraphics[trim=415 439.4 612.5 245.6, clip, width=\linewidth, height=0.67\linewidth]{figures/red_square/iui3/seathru_clear.png}};
                \draw[red,thick] (img.south west) rectangle (img.north east); 
            \end{tikzpicture}
        \end{minipage}
    \end{subfigure}
    \par
    \vspace{3mm}
    \begin{subfigure}[b]{.196\linewidth}
        \centering
        \begin{tikzpicture}
            \node[anchor=south west, inner sep=0pt] (img) at (0,0)
                {\includegraphics[width=\linewidth, height=0.67\linewidth]{new_figures/red_square/japan/gt.png}};
            \begin{scope}[x={(img.south east)},y={(img.north west)}]         \draw[red,thick] (0.62,0.8) rectangle (0.82,1.0);  
            \node [anchor=north west, white, align=left] at (0,1) {\fontsize{4}{4}\selectfont GT};
            \end{scope}
            
        \end{tikzpicture}
        \begin{minipage}{.49\linewidth}
            \includegraphics[width=\linewidth, height=0.67\linewidth]{new_figures/red_square/japan/depth_japan_gt.png}
        \end{minipage}
        \begin{minipage}{.47\linewidth}
            \begin{tikzpicture}
                \node[inner sep=0pt] (img) at (0,0)
                    {\includegraphics[trim=849.4 729.6 246.6 0, clip, width=\linewidth, height=0.67\linewidth]{new_figures/red_square/japan/gt.png}};
                \draw[red,thick] (img.south west) rectangle (img.north east); 
            \end{tikzpicture}
        \end{minipage}
    \end{subfigure}
    \begin{subfigure}[b]{.196\linewidth}
        \centering
        \begin{tikzpicture}
            \node[anchor=south west, inner sep=0pt] (img) at (0,0)
                {\includegraphics[width=\linewidth, height=0.67\linewidth]{new_figures/red_square/japan/ours.png}};
           \begin{scope}[x={(img.south east)},y={(img.north west)}]
        \draw[red,thick] (0.62,0.8) rectangle (0.82,1.0);
        \node [anchor=north west, white, align=left] at (0,1) {\fontsize{4}{4}\selectfont Ours};
        \node [anchor=north east, white, align=right] at (1,1) {\fontsize{4}{4}\selectfont PSNR=25.20};
    \end{scope}
        \end{tikzpicture}
        \begin{minipage}{.49\linewidth}
            \includegraphics[width=\linewidth, height=0.67\linewidth]{new_figures/red_square/japan/japan_depth.png}
        \end{minipage}
        \begin{minipage}{.47\linewidth}
            \begin{tikzpicture}
                \node[inner sep=0pt] (img) at (0,0)
                    {\includegraphics[trim=849.4 729.6 246.6 0, clip, width=\linewidth, height=0.67\linewidth]{new_figures/red_square/japan/ours.png}};
                \draw[red,thick] (img.south west) rectangle (img.north east); 
            \end{tikzpicture}
        \end{minipage}
    \end{subfigure}
    \begin{subfigure}[b]{.196\linewidth}
        \centering
        \begin{tikzpicture}
            \node[anchor=south west, inner sep=0pt] (img) at (0,0)
                {\includegraphics[width=\linewidth, height=0.67\linewidth]{new_figures/red_square/japan/seathru.png}};
            \begin{scope}[x={(img.south east)},y={(img.north west)}]         \draw[red,thick] (0.62,0.8) rectangle (0.82,1.0);     
            \node [anchor=north west, white, align=left] at (0,1) {\fontsize{4}{4}\selectfont SeaThru-NeRF};
            \node [anchor=north east, white, align=right] at (1,1) {\fontsize{4}{4}\selectfont PSNR=22.12};
            \end{scope}
        \end{tikzpicture}
        \begin{minipage}{.49\linewidth}
            \includegraphics[width=\linewidth, height=0.67\linewidth]{figures/red_square/japan/seathru_depth.png}
        \end{minipage}
        \begin{minipage}{.47\linewidth}
            \begin{tikzpicture}
                \node[inner sep=0pt] (img) at (0,0)
                    {\includegraphics[trim=849.4 729.6 246.6 0, clip, width=\linewidth, height=0.67\linewidth]{new_figures/red_square/japan/seathru.png}};
                \draw[red,thick] (img.south west) rectangle (img.north east); 
            \end{tikzpicture}
        \end{minipage}
    \end{subfigure}
    \begin{subfigure}[b]{.196\linewidth}
        \centering
        \begin{tikzpicture}
            \node[anchor=south west, inner sep=0pt] (img) at (0,0)
                {\includegraphics[width=\linewidth, height=0.67\linewidth]{new_figures/red_square/japan/ours_clear.png}};
            \begin{scope}[x={(img.south east)},y={(img.north west)}]         \draw[red,thick] (0.62,0.8) rectangle (0.82,1.0);
            \node [anchor=north west, white, align=left] at (0,1) {\fontsize{4}{4}\selectfont Our Restoration};
            \end{scope}
        \end{tikzpicture}
        \begin{minipage}{.49\linewidth}
            \includegraphics[width=\linewidth, height=0.67\linewidth]{new_figures/red_square/japan/ours_medium.png}
        \end{minipage}
        \begin{minipage}{.47\linewidth}
            \begin{tikzpicture}
                \node[inner sep=0pt] (img) at (0,0)
                    {\includegraphics[trim=849.4 729.6 246.6 0, clip, width=\linewidth, height=0.67\linewidth]{new_figures/red_square/japan/ours_clear.png}};
                \draw[red,thick] (img.south west) rectangle (img.north east); 
            \end{tikzpicture}
        \end{minipage}
    \end{subfigure}
    \begin{subfigure}[b]{.196\linewidth}
        \centering
        \begin{tikzpicture}
            \node[anchor=south west, inner sep=0pt] (img) at (0,0)
                {\includegraphics[width=\linewidth, height=0.67\linewidth]{figures/red_square/japan/seathru_clear.png}};
            \begin{scope}[x={(img.south east)},y={(img.north west)}]         \draw[red,thick] (0.62,0.8) rectangle (0.82,1.0);
            \node [anchor=north west, white, align=left] at (0,1) {\fontsize{4}{4}\selectfont SeaThru-NeRF's Restoration};
            \end{scope}
        \end{tikzpicture}
        \begin{minipage}{.49\linewidth}
            \includegraphics[width=\linewidth, height=0.67\linewidth]{figures/red_square/japan/seathru_medium.png}
        \end{minipage}
        \begin{minipage}{.47\linewidth}
            \begin{tikzpicture}
                \node[inner sep=0pt] (img) at (0,0)
                    {\includegraphics[trim=849.4 729.6 246.6 0, clip, width=\linewidth, height=0.67\linewidth]{figures/red_square/japan/seathru_clear.png}};
                \draw[red,thick] (img.south west) rectangle (img.north east); 
            \end{tikzpicture}
        \end{minipage}
    \end{subfigure}
    \par
    \vspace{3mm}
    \begin{subfigure}[b]{.196\linewidth}
        \centering
        \begin{tikzpicture}
            \node[anchor=south west, inner sep=0pt] (img) at (0,0)
                {\includegraphics[width=\linewidth, height=0.67\linewidth]{new_figures/red_square/panama/gt.png}};
            \begin{scope}[x={(img.south east)},y={(img.north west)}]         \draw[red,thick] (0.75,0.8) rectangle (0.95,1.0);  
            \node [anchor=north west, white, align=left] at (0,1) {\fontsize{4}{4}\selectfont GT};
            \end{scope}
            
        \end{tikzpicture}
        \begin{minipage}{.49\linewidth}
            \includegraphics[width=\linewidth, height=0.67\linewidth]{new_figures/red_square/panama/depth_panama_gt.png}
        \end{minipage}
        \begin{minipage}{.47\linewidth}
            \begin{tikzpicture}
                \node[inner sep=0pt] (img) at (0,0)
                    {\includegraphics[trim=1329 944.8 88.75 0, clip, width=\linewidth, height=0.67\linewidth]{new_figures/red_square/panama/gt.png}};
                \draw[red,thick] (img.south west) rectangle (img.north east); 
            \end{tikzpicture}
        \end{minipage}
    \end{subfigure}
    \begin{subfigure}[b]{.196\linewidth}
        \centering
        \begin{tikzpicture}
            \node[anchor=south west, inner sep=0pt] (img) at (0,0)
                {\includegraphics[width=\linewidth, height=0.67\linewidth]{new_figures/red_square/panama/ours.png}};
           \begin{scope}[x={(img.south east)},y={(img.north west)}]
        \draw[red,thick] (0.75,0.8) rectangle (0.95,1.0);
        \node [anchor=north west, white, align=left] at (0,1) {\fontsize{4}{4}\selectfont Ours};
        \node [anchor=north east, white, align=right] at (1,1) {\fontsize{4}{4}\selectfont PSNR=31.49};
    \end{scope}
        \end{tikzpicture}
        \begin{minipage}{.49\linewidth}
            \includegraphics[width=\linewidth, height=0.67\linewidth]{new_figures/red_square/panama/panama_depth.png}
        \end{minipage}
        \begin{minipage}{.47\linewidth}
            \begin{tikzpicture}
                \node[inner sep=0pt] (img) at (0,0)
                    {\includegraphics[trim=1329 944.8 88.75 0, clip, width=\linewidth, height=0.67\linewidth]{new_figures/red_square/panama/ours.png}};
                \draw[red,thick] (img.south west) rectangle (img.north east); 
            \end{tikzpicture}
        \end{minipage}
    \end{subfigure}
    \begin{subfigure}[b]{.196\linewidth}
        \centering
        \begin{tikzpicture}
            \node[anchor=south west, inner sep=0pt] (img) at (0,0)
                {\includegraphics[width=\linewidth, height=0.67\linewidth]{new_figures/red_square/panama/seathru.png}};
            \begin{scope}[x={(img.south east)},y={(img.north west)}]         \draw[red,thick] (0.75,0.8) rectangle (0.95,1.0);     
            \node [anchor=north west, white, align=left] at (0,1) {\fontsize{4}{4}\selectfont SeaThru-NeRF};
            \node [anchor=north east, white, align=right] at (1,1) {\fontsize{4}{4}\selectfont PSNR=28.61};
            \end{scope}
        \end{tikzpicture}
        \begin{minipage}{.49\linewidth}
            \includegraphics[width=\linewidth, height=0.67\linewidth]{figures/red_square/panama/seathru_depth.png}
        \end{minipage}
        \begin{minipage}{.47\linewidth}
            \begin{tikzpicture}
                \node[inner sep=0pt] (img) at (0,0)
                    {\includegraphics[trim=1329 944.8 88.75 0, clip, width=\linewidth, height=0.67\linewidth]{new_figures/red_square/panama/seathru.png}};
                \draw[red,thick] (img.south west) rectangle (img.north east); 
            \end{tikzpicture}
        \end{minipage}
    \end{subfigure}
    \begin{subfigure}[b]{.196\linewidth}
        \centering
        \begin{tikzpicture}
            \node[anchor=south west, inner sep=0pt] (img) at (0,0)
                {\includegraphics[width=\linewidth, height=0.67\linewidth]{new_figures/red_square/panama/ours_clear.png}};
            \begin{scope}[x={(img.south east)},y={(img.north west)}]         \draw[red,thick] (0.75,0.8) rectangle (0.95,1.0);
            \node [anchor=north west, white, align=left] at (0,1) {\fontsize{4}{4}\selectfont Our Restoration};
            \end{scope}
        \end{tikzpicture}
        \begin{minipage}{.49\linewidth}
            \includegraphics[width=\linewidth, height=0.67\linewidth]{new_figures/red_square/panama/ours_medium.png}
        \end{minipage}
        \begin{minipage}{.47\linewidth}
            \begin{tikzpicture}
                \node[inner sep=0pt] (img) at (0,0)
                    {\includegraphics[trim=1329 944.8 88.75 0, clip, width=\linewidth, height=0.67\linewidth]{new_figures/red_square/panama/ours_clear.png}};
                \draw[red,thick] (img.south west) rectangle (img.north east); 
            \end{tikzpicture}
        \end{minipage}
    \end{subfigure}
    \begin{subfigure}[b]{.196\linewidth}
        \centering
        \begin{tikzpicture}
            \node[anchor=south west, inner sep=0pt] (img) at (0,0)
                {\includegraphics[width=\linewidth, height=0.67\linewidth]{figures/red_square/panama/seathru_clear.png}};
            \begin{scope}[x={(img.south east)},y={(img.north west)}]         \draw[red,thick] (0.75,0.8) rectangle (0.95,1.0);
            \node [anchor=north west, white, align=left] at (0,1) {\fontsize{4}{4}\selectfont SeaThru-NeRF's Restoration};
            \end{scope}
        \end{tikzpicture}
        \begin{minipage}{.49\linewidth}
            \includegraphics[width=\linewidth, height=0.67\linewidth]{figures/red_square/panama/seathru_medium.png}
        \end{minipage}
        \begin{minipage}{.47\linewidth}
            \begin{tikzpicture}
                \node[inner sep=0pt] (img) at (0,0)
                    {\includegraphics[trim=1329 944.8 88.75 0, clip, width=\linewidth, height=0.67\linewidth]{figures/red_square/panama/seathru_clear.png}};
                \draw[red,thick] (img.south west) rectangle (img.north east); 
            \end{tikzpicture}
        \end{minipage}
    \end{subfigure}

    \caption{
    Underwater scene rendering in the 'IUI3 Red Sea' scene, 'Japanese Gardens Red Sea' scene and 'Panama' scene. 
    We compare our method with SeaThru-NeRF by showing both the full image and the rendering without the medium.
    Furthermore, under each image, we show the depth maps (for GT the depth map from pre-trained model \cite{depth_anything_v2}, and highlighted region from the image. For Restoration, we further show the rendered medium without rendering objects.
    Our method achieves better rendering quality and preserves finer distant geometric details while reducing the amount of floaters.
    }
    \label{fig:red_squre}
\end{figure*}
\begin{figure*}[ht!]
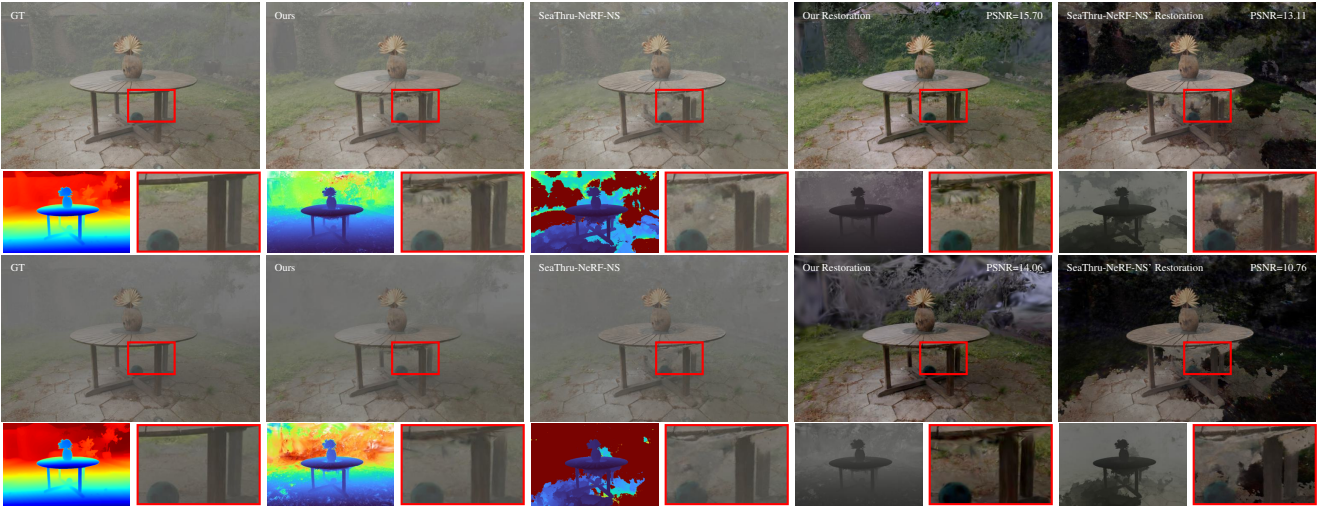

    \centering
    \begin{subfigure}[b]{.196\linewidth}
        \centering
        \begin{tikzpicture}
            \node[anchor=south west, inner sep=0pt] (img) at (0,0)
                {\includegraphics[width=\linewidth, height=0.645\linewidth]{new_figures/simulated/easy/gt.png}};
            \begin{scope}[x={(img.south east)},y={(img.north west)}]         \draw[red,thick] (0.49,0.286) rectangle (0.67,0.476);  
            \node [anchor=north west, white, align=left] at (0,1) {\fontsize{4}{4}\selectfont GT};
            \end{scope}
            
        \end{tikzpicture}
        \begin{minipage}{.49\linewidth}
            \includegraphics[width=\linewidth, height=0.645\linewidth]{new_figures/simulated/easy/gt_depth.png}
        \end{minipage}
        \begin{minipage}{.47\linewidth}
            \begin{tikzpicture}
                \node[inner sep=0pt] (img) at (0,0)
                    {\includegraphics[trim=630 240 425 440, clip, width=\linewidth, height=0.645\linewidth]{new_figures/simulated/easy/gt.png}};
                \draw[red,thick] (img.south west) rectangle (img.north east); 
            \end{tikzpicture}
        \end{minipage}
    \end{subfigure}
    \begin{subfigure}[b]{.196\linewidth}
        \centering
        \begin{tikzpicture}
            \node[anchor=south west, inner sep=0pt] (img) at (0,0)
                {\includegraphics[width=\linewidth, height=0.645\linewidth]{new_figures/simulated/easy/ours.png}};
           \begin{scope}[x={(img.south east)},y={(img.north west)}]
        \draw[red,thick] (0.49,0.286) rectangle (0.67,0.476);
        \node [anchor=north west, white, align=left] at (0,1) {\fontsize{4}{4}\selectfont Ours};
    \end{scope}
        \end{tikzpicture}
        \begin{minipage}{.49\linewidth}
            \includegraphics[width=\linewidth, height=0.645\linewidth]{new_figures/simulated/easy/ours_depth.png}
        \end{minipage}
        \begin{minipage}{.47\linewidth}
            \begin{tikzpicture}
                \node[inner sep=0pt] (img) at (0,0)
                    {\includegraphics[trim=630 240 425 440, clip, width=\linewidth, height=0.645\linewidth]{new_figures/simulated/easy/ours.png}};
                \draw[red,thick] (img.south west) rectangle (img.north east); 
            \end{tikzpicture}
        \end{minipage}
    \end{subfigure}
    \begin{subfigure}[b]{.196\linewidth}
        \centering
        \begin{tikzpicture}
            \node[anchor=south west, inner sep=0pt] (img) at (0,0)
                {\includegraphics[width=\linewidth, height=0.645\linewidth]{new_figures/simulated/easy/seathru.png}};
            \begin{scope}[x={(img.south east)},y={(img.north west)}]         \draw[red,thick] (0.49,0.286) rectangle (0.67,0.476);     
            \node [anchor=north west, white, align=left] at (0,1) {\fontsize{4}{4}\selectfont SeaThru-NeRF-NS};
            \end{scope}
        \end{tikzpicture}
        \begin{minipage}{.49\linewidth}
            \includegraphics[width=\linewidth, height=0.645\linewidth]{new_figures/simulated/easy/seathru_depth.png}
        \end{minipage}
        \begin{minipage}{.47\linewidth}
            \begin{tikzpicture}
                \node[inner sep=0pt] (img) at (0,0)
                    {\includegraphics[trim=630 240 425 440, clip, width=\linewidth, height=0.645\linewidth]{new_figures/simulated/easy/seathru.png}};
                \draw[red,thick] (img.south west) rectangle (img.north east); 
            \end{tikzpicture}
        \end{minipage}
    \end{subfigure}
    \begin{subfigure}[b]{.196\linewidth}
        \centering
        \begin{tikzpicture}
            \node[anchor=south west, inner sep=0pt] (img) at (0,0)
                {\includegraphics[width=\linewidth, height=0.645\linewidth]{new_figures/simulated/easy/ours_clear.png}};
            \begin{scope}[x={(img.south east)},y={(img.north west)}]         \draw[red,thick] (0.49,0.286) rectangle (0.67,0.476);
            \node [anchor=north west, white, align=left] at (0,1) {\fontsize{4}{4}\selectfont Our Restoration};
            \node [anchor=north east, white, align=right] at (1,1) {\fontsize{4}{4}\selectfont PSNR=15.70};
            \end{scope}
        \end{tikzpicture}
        \begin{minipage}{.49\linewidth}
            \includegraphics[width=\linewidth, height=0.645\linewidth]{new_figures/simulated/easy/ours_medium.png}
        \end{minipage}
        \begin{minipage}{.47\linewidth}
            \begin{tikzpicture}
                \node[inner sep=0pt] (img) at (0,0)
                    {\includegraphics[trim=630 240 425 440, clip, width=\linewidth, height=0.645\linewidth]{new_figures/simulated/easy/ours_clear.png}};
                \draw[red,thick] (img.south west) rectangle (img.north east); 
            \end{tikzpicture}
        \end{minipage}
    \end{subfigure}
    \begin{subfigure}[b]{.196\linewidth}
        \centering
        \begin{tikzpicture}
            \node[anchor=south west, inner sep=0pt] (img) at (0,0)
                {\includegraphics[width=\linewidth, height=0.645\linewidth]{new_figures/simulated/easy/seathru_clear.png}};
            \begin{scope}[x={(img.south east)},y={(img.north west)}]         \draw[red,thick] (0.49,0.286) rectangle (0.67,0.476);
            \node [anchor=north west, white, align=left] at (0,1) {\fontsize{4}{4}\selectfont SeaThru-NeRF-NS' Restoration};
            \node [anchor=north east, white, align=right] at (1,1) {\fontsize{4}{4}\selectfont PSNR=13.11};
            \end{scope}
        \end{tikzpicture}
        \begin{minipage}{.49\linewidth}
            \includegraphics[width=\linewidth, height=0.645\linewidth]{new_figures/simulated/easy/seathru_medium.png}
        \end{minipage}
        \begin{minipage}{.47\linewidth}
            \begin{tikzpicture}
                \node[inner sep=0pt] (img) at (0,0)
                    {\includegraphics[trim=630 240 425 440, clip, width=\linewidth, height=0.645\linewidth]{new_figures/simulated/easy/seathru_clear.png}};
                \draw[red,thick] (img.south west) rectangle (img.north east); 
            \end{tikzpicture}
        \end{minipage}
    \end{subfigure}
    \par
    \vspace{3mm}
    \begin{subfigure}[b]{.196\linewidth}
        \centering
        \begin{tikzpicture}
            \node[anchor=south west, inner sep=0pt] (img) at (0,0)
                {\includegraphics[width=\linewidth, height=0.645\linewidth]{new_figures/simulated/hard/gt.png}};
            \begin{scope}[x={(img.south east)},y={(img.north west)}]         \draw[red,thick] (0.49,0.286) rectangle (0.67,0.476);  
            \node [anchor=north west, white, align=left] at (0,1) {\fontsize{4}{4}\selectfont GT};
            \end{scope}
            
        \end{tikzpicture}
        \begin{minipage}{.49\linewidth}
            \includegraphics[width=\linewidth, height=0.645\linewidth]{new_figures/simulated/hard/depth_hard.png}
        \end{minipage}
        \begin{minipage}{.47\linewidth}
            \begin{tikzpicture}
                \node[inner sep=0pt] (img) at (0,0)
                    {\includegraphics[trim=630 240 425 440, clip, width=\linewidth, height=0.645\linewidth]{new_figures/simulated/hard/gt.png}};
                \draw[red,thick] (img.south west) rectangle (img.north east); 
            \end{tikzpicture}
        \end{minipage}
    \end{subfigure}
    \begin{subfigure}[b]{.196\linewidth}
        \centering
        \begin{tikzpicture}
            \node[anchor=south west, inner sep=0pt] (img) at (0,0)
                {\includegraphics[width=\linewidth, height=0.645\linewidth]{new_figures/simulated/hard/ours.png}};
           \begin{scope}[x={(img.south east)},y={(img.north west)}]
        \draw[red,thick] (0.49,0.286) rectangle (0.67,0.476);
        \node [anchor=north west, white, align=left] at (0,1) {\fontsize{4}{4}\selectfont Ours};
    \end{scope}
        \end{tikzpicture}
        \begin{minipage}{.49\linewidth}
            \includegraphics[width=\linewidth, height=0.645\linewidth]{new_figures/simulated/hard/ours_depth.png}
        \end{minipage}
        \begin{minipage}{.47\linewidth}
            \begin{tikzpicture}
                \node[inner sep=0pt] (img) at (0,0)
                    {\includegraphics[trim=630 240 425 440, clip, width=\linewidth, height=0.645\linewidth]{new_figures/simulated/hard/ours.png}};
                \draw[red,thick] (img.south west) rectangle (img.north east); 
            \end{tikzpicture}
        \end{minipage}
    \end{subfigure}
    \begin{subfigure}[b]{.196\linewidth}
        \centering
        \begin{tikzpicture}
            \node[anchor=south west, inner sep=0pt] (img) at (0,0)
                {\includegraphics[width=\linewidth, height=0.645\linewidth]{new_figures/simulated/hard/seathru.png}};
            \begin{scope}[x={(img.south east)},y={(img.north west)}]         \draw[red,thick] (0.49,0.286) rectangle (0.67,0.476);     
            \node [anchor=north west, white, align=left] at (0,1) {\fontsize{4}{4}\selectfont SeaThru-NeRF-NS};
            \end{scope}
        \end{tikzpicture}
        \begin{minipage}{.49\linewidth}
            \includegraphics[width=\linewidth, height=0.645\linewidth]{new_figures/simulated/hard/seathru_depth.png}
        \end{minipage}
        \begin{minipage}{.47\linewidth}
            \begin{tikzpicture}
                \node[inner sep=0pt] (img) at (0,0)
                    {\includegraphics[trim=630 240 425 440, clip, width=\linewidth, height=0.645\linewidth]{new_figures/simulated/hard/seathru.png}};
                \draw[red,thick] (img.south west) rectangle (img.north east); 
            \end{tikzpicture}
        \end{minipage}
    \end{subfigure}
    \begin{subfigure}[b]{.196\linewidth}
        \centering
        \begin{tikzpicture}
            \node[anchor=south west, inner sep=0pt] (img) at (0,0)
                {\includegraphics[width=\linewidth, height=0.645\linewidth]{new_figures/simulated/hard/ours_clear.png}};
            \begin{scope}[x={(img.south east)},y={(img.north west)}]         \draw[red,thick] (0.49,0.286) rectangle (0.67,0.476);
            \node [anchor=north west, white, align=left] at (0,1) {\fontsize{4}{4}\selectfont Our Restoration};
            \node [anchor=north east, white, align=right] at (1,1) {\fontsize{4}{4}\selectfont PSNR=14.06};
            \end{scope}
        \end{tikzpicture}
        \begin{minipage}{.49\linewidth}
            \includegraphics[width=\linewidth, height=0.645\linewidth]{new_figures/simulated/hard/ours_medium.png}
        \end{minipage}
        \begin{minipage}{.47\linewidth}
            \begin{tikzpicture}
                \node[inner sep=0pt] (img) at (0,0)
                    {\includegraphics[trim=630 240 425 440, clip, width=\linewidth, height=0.645\linewidth]{new_figures/simulated/hard/ours_clear.png}};
                \draw[red,thick] (img.south west) rectangle (img.north east); 
            \end{tikzpicture}
        \end{minipage}
    \end{subfigure}
    \begin{subfigure}[b]{.196\linewidth}
        \centering
        \begin{tikzpicture}
            \node[anchor=south west, inner sep=0pt] (img) at (0,0)
                {\includegraphics[width=\linewidth, height=0.645\linewidth]{new_figures/simulated/hard/seathru_clear.png}};
            \begin{scope}[x={(img.south east)},y={(img.north west)}]         \draw[red,thick] (0.49,0.286) rectangle (0.67,0.476);
            \node [anchor=north west, white, align=left] at (0,1) {\fontsize{4}{4}\selectfont SeaThru-NeRF-NS' Restoration};
            \node [anchor=north east, white, align=right] at (1,1) {\fontsize{4}{4}\selectfont PSNR=10.76};
            \end{scope}
        \end{tikzpicture}
        \begin{minipage}{.49\linewidth}
            \includegraphics[width=\linewidth, height=0.645\linewidth]{new_figures/simulated/hard/seathru_medium.png}
        \end{minipage}
        \begin{minipage}{.47\linewidth}
            \begin{tikzpicture}
                \node[inner sep=0pt] (img) at (0,0)
                    {\includegraphics[trim=630 240 425 440, clip, width=\linewidth, height=0.645\linewidth]{new_figures/simulated/hard/seathru_clear.png}};
                \draw[red,thick] (img.south west) rectangle (img.north east); 
            \end{tikzpicture}
        \end{minipage}
    \end{subfigure}
    
    \caption{Simulated scene rendering with the easy foggy scene (upper) and hard foggy scene (lower). We compare our method with SeaThru-NeRF by showing both the full image and the rendering without the attenuation (restoration).
   Furthermore, under each image, we show the depth maps (for GT the depth map from pre-trained model \cite{depth_anything_v2}), and highlighted region from the image. For restoration, we further show the rendered medium without rendering objects.
    Our results exhibit better restoring quality and reasonable depth map compared to SeaThru-NeRF-NS' results.
    }
    \label{fig:red_squre_simulated}
\end{figure*}
\begin{table*}[!ht]
    \centering
    \caption{\textbf{Quantitative evaluation on the SeaThru-NeRF dataset.} We show PSNR$\uparrow$, SSIM$\uparrow$, LPIPS$\downarrow$, Avg. FPS$\uparrow$, and Avg. Training Time$\downarrow$. The \colorbox{lightred}{first}, \colorbox{lightorange}{second}, and \colorbox{lightyellow}{third} values are highlighted.}
    \label{tab:seathru_compare}
    \small  %
    \setlength{\tabcolsep}{4pt} %
    \resizebox{\textwidth}{!}{%
    \begin{tabular}{l|ccc|ccc|ccc|ccc|c|c}
        \textbf{Dataset/Metric} & \multicolumn{3}{c|}{\textbf{IUI3 Red Sea}} & \multicolumn{3}{c|}{\textbf{Curaçao}} & \multicolumn{3}{c|}{\textbf{J.G. Red Sea}} & \multicolumn{3}{c|}{\textbf{Panama}} & Avg. & Avg. \\
        \textbf{Method} & PSNR & SSIM & LPIPS & PSNR & SSIM & LPIPS & PSNR & SSIM & LPIPS & PSNR & SSIM & LPIPS & FPS & Time\\
        \hline
        SeaThru-NeRF & 27.310& 0.861 & 0.221 & 29.305& 0.898 & 0.214 & 22.116& 0.802 & 0.232 & 28.608& 0.868 & 0.202 & 0.07 & 10h \\
        SeaThru-NeRF-NS & \cellcolor{lightyellow}27.651& \cellcolor{lightyellow}0.885& \cellcolor{lightorange}0.133& \cellcolor{lightorange}30.609& \cellcolor{lightyellow}0.928& \cellcolor{lightyellow}0.159& \cellcolor{lightorange}23.542& \cellcolor{lightorange}0.889& \cellcolor{lightorange}0.116& \cellcolor{lightorange}31.857& \cellcolor{lightyellow}0.948& \cellcolor{lightorange}0.071& 0.9& 2h\\
        ZipNeRF & \cellcolor{lightorange}29.350 & \cellcolor{lightorange}0.899 & \cellcolor{lightred}0.106 & \cellcolor{lightyellow}29.925 & \cellcolor{lightorange}0.938 & \cellcolor{lightorange}0.124 & \cellcolor{lightyellow}23.446 & \cellcolor{lightyellow}0.883 & \cellcolor{lightyellow}0.136 & \cellcolor{lightred}32.335 & \cellcolor{lightred}0.956 & \cellcolor{lightred}0.064 & 0.9& 6h \\
        3D Gauss. & 22.980 & 0.843& 0.2458& 28.313& 0.873& 0.221& 21.493 & 0.854& 0.216& 29.200& 0.893& 0.152& 412.1& 17.4min\\
        Ours & \cellcolor{lightred}29.386& \cellcolor{lightred}0.910& \cellcolor{lightyellow}0.180& \cellcolor{lightred}32.672& \cellcolor{lightred}0.957& \cellcolor{lightred}0.110& \cellcolor{lightred}25.201& \cellcolor{lightred}0.904& \cellcolor{lightred}0.113& \cellcolor{lightyellow}31.490& \cellcolor{lightorange}0.948& \cellcolor{lightyellow}0.075& 41.8& 9.4min\\
    \end{tabular}%
    }
\end{table*}

\textbf{Implementation Details}: Our implementation is based on the reimplemented version of 3DGS released by NeRF-Studio \cite{ye2024gsplat}. Following \cite{ye2024absgs, yu2024gaussian}, we accumulate the norms of the individual pixel gradients of $\mu_i$ for primitive densification.
 For the medium encoding, we use a spherical harmonic encoding \cite{verbin2022refnerf} and a MLP with 2 linear layers with 128 hidden units and Sigmoid activation, followed by Sigmoid activation for $c^{\text{med}}$ and Softplus activation for $\sigma^{\text{attn}}$ and $\sigma^{\text{bs}}$. Upon each densification and pruning step of the 3DGS, the moving averages in the Adam optimizer of the medium encoding are reset to ensure the independence of subsequent iterations. We reset opacity to 0.5 every 500 training steps and prune gaussians with opacity below 0.5 every 100 training steps.
 The culling and reset thresholds \cite{kerbl3Dgaussians} are set to 0.5 to encourage Gaussians to model opaque objects.
\\
\textbf{SeaThru-NeRF Dataset}: SeaThru-NeRF Dataset released by \cite{levy2023seathrunerf} contains real-world scenes acquired from four different scenes in sea: IUI3 Red Sea, Curaçao, Japanese Gardens Red Sea, and Panama. There are 29, 20, 20 and 18 images respectively, among which 25, 17, 17 and 15 images are used for training and the rest 4, 3, 3 and 3 are used for validation. The dataset encompasses a variety of water and imaging conditions. The images were captured in RAW format using a Nikon D850 SLR camera housed in a Nauticam underwater casing with a dome port, which helped prevent refractions that could disrupt the pinhole model. These images were then downsampled to an approximate resolution of 900 × 1400. Prior to processing, the input linear images underwent white balancing with a $0.5\%$ clipping per channel to eliminate extreme noise pixels. Lastly, COLMAP \cite{Colmap} was employed to determine the camera poses and correct lens distortions inherent.
\\
\textbf{Simulated Dataset:}
To further evaluate the performance of the proposed method, we take a standard NeRF dataset - the Garden scene from the Mip-NeRF 360 dataset \cite{barron2022mipnerf360} - and added fog to it to simulate the presence of medium. We used 3DGS to extract the depth maps. These maps were then utilized to create scenarios simulating both underwater and foggy conditions. In line with method Eq. (\ref{eq:revised}), we utilized the following parameters to simulate easy foggy scenario: $\beta ^{D} = [0.6, 0.6, 0.6]$, $\beta ^{B} = [0.6, 0.6, 0.6]$, and $B^{\infty} = [0.5, 0.5, 0.5]$.  The parameters for the hard foggy case are: $\beta^{D} = [0.8, 0.8, 0.8]$, $\beta^{B} = [0.6, 0.6, 0.6]$, and $B^{\infty} = [0.5, 0.5, 0.5]$.
\\
\textbf{Baseline methods:}
All methods were trianed on the same set of white-balanced images. For rendering scenes with the medium, we compare several NeRF techniques: SeaThru-NeRF \cite{levy2023seathrunerf}, the reimplementation of SeaThru-NeRF released on NeRF-Studio (SeaThru-NeRF-NS) \cite{setinek2023seathru}, Zip-NeRF \cite{barron2023zipnerf} and 3DGS \cite{kerbl3Dgaussians}. For each baseline method, we use the PSNR, SSIM, and LPIPS \cite{zhang2018lpips} metrics to compare rendering quality. We present the alpha blending of depth as the depth map and the rendering without medium to demonstrate the ability to decouple the medium and the object for SeaThru-NeRF and our method. We also calculate the FPS and total training time using the same RTX 4080 GPU to illustrate the speed difference between baselines and our method. All reported results are averaged over three runs.

\subsection{Results}
First, we evaluated the performance of our method using the standard benchmark dataset, the SeaThru-NeRF Dataset. Table \ref{tab:seathru_compare} compares PSNR, SSIM, LPIPS, average FPS and average training time with the baseline methods across the validation sets of four scenes. Our method demonstrates its superiority in the majority of cases and efficiency on both rendering and training. 'Panama' is a special case where there is little medium and the medium properties stays the same across different ray directions. Therefore, ZipNeRF can reconstruct it well. However, ZipNeRF training takes orders of magnitude more time than our method and does not offer real-time rendering. 

Fig. \ref{fig:curasao_baselines} shows that the mainstream 3DGS and NeRF approachs are short at reconstruction on back-scattering media. 3DGS prunes the Gaussians with low opacity, leaving dense and muddy cloud-like primitives to fit the medium, which causes artifacts in the novel views. Zip-NeRF struggles to model the geometrical surface, leading to an unreal scene reconstructed with little media left.

Fig. \ref{fig:red_squre} demonstrates that in the 'IUI3 Red Sea', the 'Japanese Gardens Red Sea' and 'Panama' scenes, 
our method delivers superior quality and more effectively separates medium and object than the SeaThru-NeRF, especially in deeper and more complex scenes (as highlighted in the red square). Additionally, our depth map reveals much finer details compared to SeaThru-NeRF, which struggles to produce a reasonable depth map at greater distances, as indicated by the red color in the upper right corner of the depth map. We also achieve higher PSNR values in both scenes. The same advantages are observed in simulated scenes, where our method renders better details (indicated by the red square) than the SeaThru-NeRF in both easy and hard foggy scenes as depicted in Fig. \ref{fig:red_squre_simulated}. Our rendering without medium and depth maps significantly outperform those from the SeaThru-NeRF, especially in scenes that are farther from the camera. While our method's predictions may appear blurry and the object map unclear in the upper right corner for the hard foggy scene, the results from SeaThru-NeRF are considerably worse. The restoration quality comparison presented in Table \ref{tab:restore_compare} further quantitatively demonstrates the superiority of our method in simulated scenes. Overall, our method surpasses SeaThru-NeRF in both underwater and simulated scenes.

\begin{table}[!htbp]
    \centering
    \caption{\textbf{Restoration Performance.} 
    (PSNR$\uparrow$/SSIM$\uparrow$/LPIPS$\downarrow$)
    \label{tab:restore_compare}
    }
    \small  %
    \resizebox{\columnwidth}{!}{
        \begin{tabular}{l|ccc|ccc}
            \textbf{Dataset/Metric} & \multicolumn{3}{c|}{\textbf{Foggy-Easy}} & \multicolumn{3}{c}{\textbf{Foggy-Hard}} \\
            \textbf{Method} & PSNR & SSIM & LPIPS & PSNR & SSIM & LPIPS \\
            \hline
            SeaThru-NeRF-NS & 13.11 & 0.32 & 0.58 & 10.76 & 0.29 & 0.63\\
            Ours & \cellcolor{lightred}15.70 & \cellcolor{lightred}0.37 & \cellcolor{lightred}0.56 & \cellcolor{lightred}14.06 & \cellcolor{lightred}0.45 & \cellcolor{lightred}0.54\\
        \end{tabular}%
    }
\end{table}
\begin{figure*}[ht!]
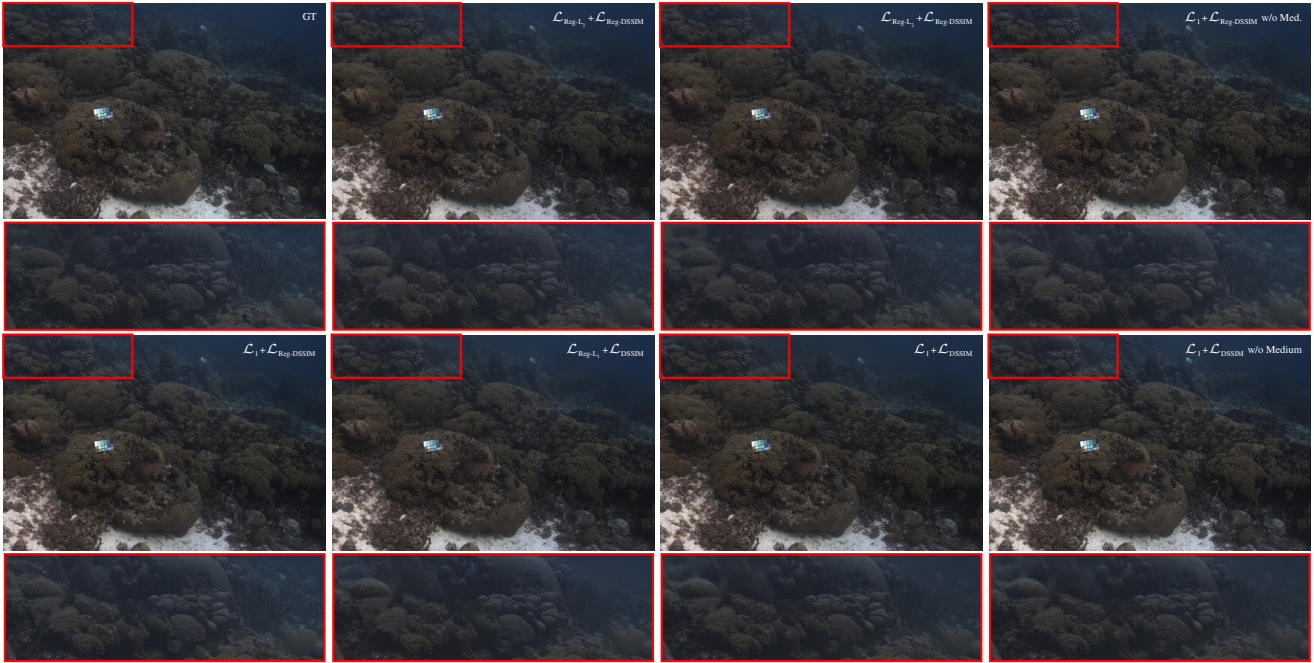

    \centering
    \begin{subfigure}[b]{.194\linewidth}
        \centering
        \begin{tikzpicture}
            \node[anchor=south west, inner sep=0pt] (img) at (0,0)
                {\includegraphics[width=\linewidth, height=0.67\linewidth]{new_figures/ablation/gt.png}};
            \begin{scope}[x={(img.south east)},y={(img.north west)}]         \draw[red,thick] (0,0.8) rectangle (0.4,1);
            \node [anchor=north east, white, align=right] at (1,1) {\fontsize{4}{4}\selectfont GT};
            \end{scope}
        \end{tikzpicture}
        \begin{minipage}{.99\linewidth}
            \begin{tikzpicture}
                \node[inner sep=0pt] (img) at (0,0)
                    {\includegraphics[trim=0 944.8 1065 0, clip, width=\linewidth]{new_figures/ablation/gt.png}};
                \draw[red,thick] (img.south west) rectangle (img.north east); 
            \end{tikzpicture}
        \end{minipage}
    \end{subfigure}
    \begin{subfigure}[b]{.194\linewidth}
        \centering
        \begin{tikzpicture}
            \node[anchor=south west, inner sep=0pt] (img) at (0,0)
                {\includegraphics[width=\linewidth, height=0.67\linewidth]{new_figures/ablation/regl2+regssim.png}};
            \begin{scope}[x={(img.south east)},y={(img.north west)}]         \draw[red,thick] (0,0.8) rectangle (0.4,1);
            \node [anchor=north east, white, align=right] at (1,1) {\fontsize{4}{4}\selectfont $\mathcal{L}_{\text{Reg-L}_{\text{2}}}$+$\mathcal{L}_{\fontsize{4}{4}\selectfont \text{Reg-DSSIM}}$};
            \end{scope}
        \end{tikzpicture}
        \begin{minipage}{.99\linewidth}
            \begin{tikzpicture}
                \node[inner sep=0pt] (img) at (0,0)
                    {\includegraphics[trim=0 944.8 1065 0, clip, width=\linewidth]{new_figures/ablation/regl2+regssim.png}};
                \draw[red,thick] (img.south west) rectangle (img.north east); 
            \end{tikzpicture}
        \end{minipage}
    \end{subfigure}
    \begin{subfigure}[b]{.194\linewidth}
        \centering
        \begin{tikzpicture}
            \node[anchor=south west, inner sep=0pt] (img) at (0,0)
                {\includegraphics[width=\linewidth, height=0.67\linewidth]{new_figures/ablation/l2+regssim.png}};
            \begin{scope}[x={(img.south east)},y={(img.north west)}]         \draw[red,thick] (0,0.8) rectangle (0.4,1);
            \node [anchor=north east, white, align=right] at (1,1) {\fontsize{4}{4}\selectfont $\mathcal{L}_{\text{2}}$+$\mathcal{L}_{\fontsize{4}{4}\selectfont \text{Reg-DSSIM}}$};
            \end{scope}
        \end{tikzpicture}
        \begin{minipage}{.99\linewidth}
            \begin{tikzpicture}
                \node[inner sep=0pt] (img) at (0,0)
                    {\includegraphics[trim=0 944.8 1065 0, clip, width=\linewidth]{new_figures/ablation/l2+regssim.png}};
                \draw[red,thick] (img.south west) rectangle (img.north east); 
            \end{tikzpicture}
        \end{minipage}
    \end{subfigure}
    \begin{subfigure}[b]{.194\linewidth}
        \centering
        \begin{tikzpicture}
            \node[anchor=south west, inner sep=0pt] (img) at (0,0)
                {\includegraphics[width=\linewidth, height=0.67\linewidth]{new_figures/ablation/l2+ssim.png}};
            \begin{scope}[x={(img.south east)},y={(img.north west)}]         \draw[red,thick] (0,0.8) rectangle (0.4,1);
            \node [anchor=north east, white, align=right] at (1,1) {\fontsize{4}{4}\selectfont $\mathcal{L}_{\fontsize{4}{4}\selectfont \text{2}}$+$\mathcal{L}_{\fontsize{4}{4}\selectfont \text{DSSIM}}$};
            \end{scope}
        \end{tikzpicture}
        \begin{minipage}{.99\linewidth}
            \begin{tikzpicture}
                \node[inner sep=0pt] (img) at (0,0)
                    {\includegraphics[trim=0 944.8 1065 0, clip, width=\linewidth]{new_figures/ablation/l2+ssim.png}};
                \draw[red,thick] (img.south west) rectangle (img.north east); 
            \end{tikzpicture}
        \end{minipage}
    \end{subfigure}
    \begin{subfigure}[b]{.194\linewidth}
        \centering
        \begin{tikzpicture}
            \node[anchor=south west, inner sep=0pt] (img) at (0,0)
                {\includegraphics[width=\linewidth, height=0.67\linewidth]{new_figures/ablation/regl2+regssim-med.png}};
            \begin{scope}[x={(img.south east)},y={(img.north west)}]         \draw[red,thick] (0,0.8) rectangle (0.4,1);
            \node [anchor=north east, white, align=right] at (1,1) {\fontsize{4}{4}\selectfont $\mathcal{L}_{\fontsize{4}{4}\selectfont \text{2}}$+$\mathcal{L}_{\fontsize{4}{4}\selectfont \text{Reg-DSSIM}}$ w/o Med.};
            \end{scope}
        \end{tikzpicture}
        \begin{minipage}{.99\linewidth}
            \begin{tikzpicture}
                \node[inner sep=0pt] (img) at (0,0)
                    {\includegraphics[trim=0 944.8 1065 0, clip, width=\linewidth]{new_figures/ablation/regl2+regssim-med.png}};
                \draw[red,thick] (img.south west) rectangle (img.north east); 
            \end{tikzpicture}
        \end{minipage}
    \end{subfigure}

    \begin{subfigure}[b]{.194\linewidth}
        \centering
        \begin{tikzpicture}
            \node[anchor=south west, inner sep=0pt] (img) at (0,0)
                {\includegraphics[width=\linewidth, height=0.67\linewidth]{new_figures/ablation/l1+ssim-med.png}};
            \begin{scope}[x={(img.south east)},y={(img.north west)}]         \draw[red,thick] (0,0.8) rectangle (0.4,1);
            \node [anchor=north east, white, align=right] at (1,1) {\fontsize{4}{4}\selectfont $\mathcal{L}_{\text{1}}$+$\mathcal{L}_{\fontsize{4}{4}\selectfont \text{DSSIM}}$ w/o Med.};
            \end{scope}
        \end{tikzpicture}
        \begin{minipage}{.99\linewidth}
            \begin{tikzpicture}
                \node[inner sep=0pt] (img) at (0,0)
                    {\includegraphics[trim=0 944.8 1065 0, clip, width=\linewidth]{new_figures/ablation/l1+ssim-med.png}};
                \draw[red,thick] (img.south west) rectangle (img.north east); 
            \end{tikzpicture}
        \end{minipage}
    \end{subfigure}
    \begin{subfigure}[b]{.194\linewidth}
        \centering
        \begin{tikzpicture}
            \node[anchor=south west, inner sep=0pt] (img) at (0,0)
                {\includegraphics[width=\linewidth, height=0.67\linewidth]{new_figures/ablation/regl1+regssim.png}};
            \begin{scope}[x={(img.south east)},y={(img.north west)}]         \draw[red,thick] (0,0.8) rectangle (0.4,1);
            \node [anchor=north east, white, align=right] at (1,1) {\fontsize{4}{4}\selectfont $\mathcal{L}_{\text{Reg-L}_{\text{1}}}$+$\mathcal{L}_{\fontsize{4}{4}\selectfont \text{Reg-DSSIM}}$};
            \end{scope}
        \end{tikzpicture}
        \begin{minipage}{.99\linewidth}
            \begin{tikzpicture}
                \node[inner sep=0pt] (img) at (0,0)
                    {\includegraphics[trim=0 944.8 1065 0, clip, width=\linewidth]{new_figures/ablation/regl1+regssim.png}};
                \draw[red,thick] (img.south west) rectangle (img.north east); 
            \end{tikzpicture}
        \end{minipage}
    \end{subfigure}
    \begin{subfigure}[b]{.194\linewidth}
        \centering
        \begin{tikzpicture}
            \node[anchor=south west, inner sep=0pt] (img) at (0,0)
                {\includegraphics[width=\linewidth, height=0.67\linewidth]{new_figures/ablation/l1+regssim.png}};
            \begin{scope}[x={(img.south east)},y={(img.north west)}]         \draw[red,thick] (0,0.8) rectangle (0.4,1);
            \node [anchor=north east, white, align=right] at (1,1) {\fontsize{4}{4}\selectfont $\mathcal{L}_{\text{1}}$+$\mathcal{L}_{\fontsize{4}{4}\selectfont \text{Reg-DSSIM}}$};
            \end{scope}
        \end{tikzpicture}
        \begin{minipage}{.99\linewidth}
            \begin{tikzpicture}
                \node[inner sep=0pt] (img) at (0,0)
                    {\includegraphics[trim=0 944.8 1065 0, clip, width=\linewidth]{new_figures/ablation/l1+regssim.png}};
                \draw[red,thick] (img.south west) rectangle (img.north east); 
            \end{tikzpicture}
        \end{minipage}
    \end{subfigure}
    \begin{subfigure}[b]{.194\linewidth}
        \centering
        \begin{tikzpicture}
            \node[anchor=south west, inner sep=0pt] (img) at (0,0)
                {\includegraphics[width=\linewidth, height=0.67\linewidth]{new_figures/ablation/l1+ssim.png}};
            \begin{scope}[x={(img.south east)},y={(img.north west)}]         \draw[red,thick] (0,0.8) rectangle (0.4,1);
            \node [anchor=north east, white, align=right] at (1,1) {\fontsize{4}{4}\selectfont $\mathcal{L}_{\fontsize{4}{4}\selectfont \text{1}}$+$\mathcal{L}_{\fontsize{4}{4}\selectfont \text{DSSIM}}$};
            \end{scope}
        \end{tikzpicture}
        \begin{minipage}{.99\linewidth}
            \begin{tikzpicture}
                \node[inner sep=0pt] (img) at (0,0)
                    {\includegraphics[trim=0 944.8 1065 0, clip, width=\linewidth]{new_figures/ablation/l1+ssim.png}};
                \draw[red,thick] (img.south west) rectangle (img.north east); 
            \end{tikzpicture}
        \end{minipage}
    \end{subfigure}
    \begin{subfigure}[b]{.194\linewidth}
        \centering
        \begin{tikzpicture}
            \node[anchor=south west, inner sep=0pt] (img) at (0,0)
                {\includegraphics[width=\linewidth, height=0.67\linewidth]{new_figures/ablation/regl1+regssim-med.png}};
            \begin{scope}[x={(img.south east)},y={(img.north west)}]         \draw[red,thick] (0,0.8) rectangle (0.4,1);
            \node [anchor=north east, white, align=right] at (1,1) {\fontsize{4}{4}\selectfont $\mathcal{L}_{\fontsize{4}{4}\selectfont \text{1}}$+$\mathcal{L}_{\fontsize{4}{4}\selectfont \text{Reg-DSSIM}}$ w/o Med.};
            \end{scope}
        \end{tikzpicture}
        \begin{minipage}{.99\linewidth}
            \begin{tikzpicture}
                \node[inner sep=0pt] (img) at (0,0)
                    {\includegraphics[trim=0 944.8 1065 0, clip, width=\linewidth]{new_figures/ablation/regl1+regssim-med.png}};
                \draw[red,thick] (img.south west) rectangle (img.north east); 
            \end{tikzpicture}
        \end{minipage}
    \end{subfigure}
    
    \caption{\textbf{Ablation Study: loss function alignment.} Our proposed $\mathcal{L}_{\text{Reg-DSSIM}}$ improves the reconstruction quality of distant details in dark areas, and the benefit is obvious even when used alone. Regularized pixel-level losses $\mathcal{L}_{\text{Reg-}\mathcal{L}_{2}}$ and $\mathcal{L}_{\text{Reg-}\mathcal{L}_{1}}$ further improve the quality of the reconstruction.
    }
    \label{fig:ablation}
    \vspace{-0.2cm}
\end{figure*}

\subsection{Ablation Study}
\label{subsec:Ablation Study}
We isolate the different contributions to show their importance. We conduct a quantitative analysis on different combination of loss functions, between pixel-wise component \{$\mathcal{L}_{1}$, $\mathcal{L}_{2}$, $\mathcal{L}_{\text{Reg-}\mathcal{L}_{1}}$, $\mathcal{L}_{\text{Reg-}\mathcal{L}_{2}}$\} and frame-wise \{$\mathcal{L}_{\text{DSSIM}}$, $\mathcal{L}_{\text{Reg-DSSIM}}$\}, as well as removing the medium effect and removing both the medium and our proposed loss function $\mathcal{L}_{\text{Reg}}$ as a reimplementation of 3DGS. These comparisons are made across validation sets for the SeaThru-NeRF dataset in Table \ref{tab:metrics_comparison}.

Fig. \ref{fig:ablation} clearly demonstrates that our frame-level $\mathcal{L}_{\text{Reg-DSSIM}}$ improves the reconstruction quality and structural similarity of distant objects in low light. When used alone, $\mathcal{L}_{\text{Reg-}\mathcal{L}_{1}}$ can also provide better details in far distance. Used with $\mathcal{L}_{\text{Reg-DSSIM}}$, pixel-level $\mathcal{L}_{\text{Reg-}\mathcal{L}_{1}}$ and $\mathcal{L}_{\text{Reg-}\mathcal{L}_{2}}$ can further enhance reconstruction quality. 
However, $\mathcal{L}_{\text{Reg-}\mathcal{L}_{2}}$ alone \cite{mildenhall2021nerf} is not sufficient to train 3DGS-based models, and $\mathcal{L}_{\text{Reg-DSSIM}}$ is required. Our proposed $\mathcal{L}_{\text{Reg}}$ shows superiority over other configurations in leading the 3DGS-based model to better fit HDR scenes and removing the medium component (basically 3DGS) significantly hurms the performance of our method, which indicates the necessity of our approach.
\begin{table}[!htbp]
\small
    \centering
    \caption{\textbf{Ablation Study} Avg. over SeaThru-NeRF Scenes}
    \label{tab:metrics_comparison}
    \vspace{-0.2cm}
    \begin{tabular}{lccc}
        \textbf{Configuration} & \textbf{PSNR$\uparrow$}& \textbf{SSIM$\uparrow$}& \textbf{LPIPS$\downarrow$}\\
        \midrule
        $\mathcal{L}_{1}$+$\mathcal{L}_{\text{DSSIM}}$ & 29.219 & 0.915 & 0.161 \\
        $\mathcal{L}_{2}$+$\mathcal{L}_{\text{DSSIM}}$ & 28.652 & 0.911 & 0.172 \\
        $\mathcal{L}_{1}$+$\mathcal{L}_{\text{Reg-DSSIM}}$ & 29.574 & 0.929 & 0.121 \\
        $\mathcal{L}_{2}$+$\mathcal{L}_{\text{Reg-DSSIM}}$ & 29.169 & 0.924 & 0.132 \\
        
        $\mathcal{L}_{\text{Reg-}\mathcal{L}_{1}}$+$\mathcal{L}_{\text{DSSIM}}$ & 28.890 & 0.916 & 0.153 \\
        $\mathcal{L}_{\text{Reg-}\mathcal{L}_{2}}$+$\mathcal{L}_{\text{DSSIM}}$ & 29.064 & 0.914 & 0.165 \\
        $\mathcal{L}_{\text{Reg-}\mathcal{L}_{1}}$+$\mathcal{L}_{\text{Reg-DSSIM}}$ & 29.603 & 0.928 & 0.123 \\
        \textbf{$\mathcal{L}_{\text{Reg-}\mathcal{L}_{2}}$+$\mathcal{L}_{\text{Reg-DSSIM}}$ (Ours)} & \cellcolor{lightred}29.687 & \cellcolor{lightred}0.930 & \cellcolor{lightred}0.119 \\
        $\mathcal{L}_{1}$+$\mathcal{L}_{\text{DSSIM}}$ w/o Medium & 29.227 & 0.915 & 0.164 \\
        $\mathcal{L}_{\text{Reg-}\mathcal{L}_{1}}$+$\mathcal{L}_{\text{Reg-DSSIM}}$ w/o Med. & 29.351 & 0.923 & 0.134 \\
        $\mathcal{L}_{\text{Reg-}\mathcal{L}_{2}}$+$\mathcal{L}_{\text{Reg-DSSIM}}$ w/o Med.& 29.353 & 0.925 & 0.130 \\

    \end{tabular}
    \vspace{-0.2cm}
\end{table}

\begin{figure}[ht!]
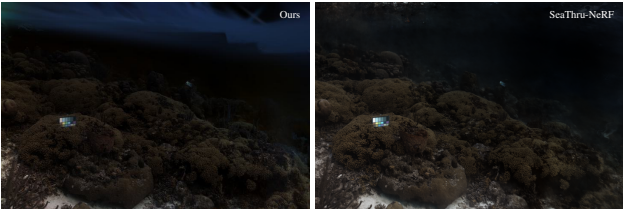

    \centering
    \begin{subfigure}[b]{.49\linewidth}
        \centering
        \begin{tikzpicture}
            \node[anchor=south west, inner sep=0pt] (img) at (0,0)
                {\includegraphics[width=\linewidth, height=0.67\linewidth]{new_figures/curasao_obj/ours.png}};
            \begin{scope}[x={(img.south east)},y={(img.north west)}]         
            \node [anchor=north east, white, align=right] at (1,1) {\fontsize{4}{4}\selectfont Ours};
            \end{scope}
            
        \end{tikzpicture}
    \end{subfigure}
    \begin{subfigure}[b]{.49\linewidth}
        \centering
        \begin{tikzpicture}
            \node[anchor=south west, inner sep=0pt] (img) at (0,0)
                {\includegraphics[width=\linewidth, height=0.67\linewidth]{figures/curasao_obj/seathru.png}};
           \begin{scope}[x={(img.south east)},y={(img.north west)}]
        
        \node [anchor=north east, white, align=right] at (1,1) {\fontsize{4}{4}\selectfont SeaThru-NeRF};
    \end{scope}
        \end{tikzpicture}
    \end{subfigure}
    
    \caption{
\textbf{Limitation: simulating distant medium with Gaussians.} Our method (\textbf{left}) models distant medium with Gaussians. SeaThru-NeRF \cite{levy2023seathrunerf} (\textbf{right}) also struggles with the background.
    }
    \label{fig:curasao_obj}
    \vspace{-0.1cm}
\end{figure}

\begin{figure}[ht!]
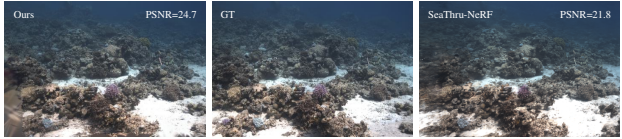

    \centering
    \begin{subfigure}[b]{.32\linewidth}
        \centering
        \begin{tikzpicture}
            \node[anchor=south west, inner sep=0pt] (img) at (0,0)
                {\includegraphics[width=\linewidth, height=0.67\linewidth]{new_figures/limitation/ours.png}};
            \begin{scope}[x={(img.south east)},y={(img.north west)}]         
            \node [anchor=north west, white, align=left] at (0,1) {\fontsize{4}{4}\selectfont Ours};
            \node [anchor=north east, white, align=right] at (1,1) {\fontsize{4}{4}\selectfont PSNR=25.2};
            \end{scope}
            
        \end{tikzpicture}
    \end{subfigure}
    \begin{subfigure}[b]{.32\linewidth}
        \centering
        \begin{tikzpicture}
            \node[anchor=south west, inner sep=0pt] (img) at (0,0)
                {\includegraphics[width=\linewidth, height=0.67\linewidth]{new_figures/limitation/gt.png}};
           \begin{scope}[x={(img.south east)},y={(img.north west)}]
        
        \node [anchor=north west, white, align=left] at (0,1) {\fontsize{4}{4}\selectfont GT};
        
    \end{scope}
        \end{tikzpicture}
    \end{subfigure}
    \begin{subfigure}[b]{.32\linewidth}
        \centering
        \begin{tikzpicture}
            \node[anchor=south west, inner sep=0pt] (img) at (0,0)
                {\includegraphics[width=\linewidth, height=0.67\linewidth]{new_figures/limitation/seathru.png}};
           \begin{scope}[x={(img.south east)},y={(img.north west)}]
        
        \node [anchor=north west, white, align=left] at (0,1) {\fontsize{4}{4}\selectfont SeaThru-NeRF};
        \node [anchor=north east, white, align=right] at (1,1) {\fontsize{4}{4}\selectfont PSNR=22.1};
    \end{scope}
        \end{tikzpicture}
    \end{subfigure}
    
    \caption{
    \textbf{Limitation: insufficient supervision.}
    Our method (\textbf{left}) has low-detail visuals in regions not sufficiently covered by training views. SeaThru-NeRF \cite{levy2023seathrunerf} (\textbf{right}) is blurry in these regions.
    }
    \label{fig:bad_case}
    \vspace{-0.1cm}
\end{figure}

\section{Limitations}
Although our method achieves good reconstruction quality, there are some limitations to consider. 
Firstly, our method, similar to NeRF-based approaches \cite{levy2023seathrunerf}, has some difficulties with distinguishing the background-like object and the medium far in the distance, as illustrated on the top of Fig. \ref{fig:curasao_baselines} and Fig. \ref{fig:curasao_obj}. However, in the foreground, our method prunes medium-role primitives well while SeaThru-NeRF cannot prevent the geometrical field from fitting the medium, resulting in wave-like artifacts.
Secondly, same as other NVS methods \cite{kerbl3Dgaussians, mildenhall2020nerf}, our method relies on the camera poses being available which might prove difficult to obtain in underwater 3D scenes. 
Thirdly, our 3DGS-based method has artifacts in regions lacking observation \cite{kerbl3Dgaussians}, which is also suffered by NeRF-based models, as illustrated in the left side of Fig. \ref{fig:bad_case} and the top part of Fig. \ref{fig:red_squre}, while the NeRF-based SeaThru-NeRF approach (right image) will introduces some blurring, distortion and interpolation.
Lastly, the restored color of the scene from the scene is not ensured to be precise (especially for the background-like object), as under the effect of medium, the color of object and the attenuation attribute are entangled during training, which is shown by Fig. \ref{fig:red_squre_simulated}.

\section{Conclusions}
In our work, we focused on the problem of underwater reconstruction, previously tackled by fully volumetric representations that are slow to train and render. Therefore, we proposed to fuse the explicit point-splatting method (3DGS) with volume rendering to achieve both fast training and real-time rendering speed.
Our method interleaves alpha compositing of splatted Gaussians with integrated ray segments passing through the scattering medium.
We have demonstrated that our method achieves state-of-the-art results while enabling real-time rendering.
Furthermore, the explicit scene representation enables disentanglement of geometry and the scattering medium.
In future work, we would like to extend our method for larger scenes with both water and fog.

{\setlength\parindent{0pt}
\vspace{1.0em}
\textbf{Acknowledgements.} This work was supported by the Czech Science Foundation (GA\v{C}R) EXPRO (grant no. 23-07973X), and by the Ministry of Education, Youth and Sports of the Czech Republic through the e-INFRA CZ (ID:90254).
Jonas Kulhanek acknowledges travel support from the European Union’s Horizon 2020 research and innovation programme under ELISE (grant no. 951847).
}

{
    \small
    \bibliographystyle{ieeenat_fullname}
    \bibliography{main}
}

\immediate\closein\imgstream

\end{document}